\documentclass[review,10pt]{elsarticle} 
\usepackage{float}
\usepackage{amsmath}
\usepackage{graphicx}
\usepackage[unicode=true, bookmarks=false]{hyperref}
\usepackage[latin9]{inputenc}
\usepackage{times} 
\usepackage[verbose,tmargin=1in,bmargin=1in,lmargin=1in,rmargin=1in]{geometry}
\usepackage{setspace}
\usepackage{multicol}
\usepackage{lineno}
\usepackage{subcaption}
\usepackage{threeparttable}
\usepackage{csquotes}
\usepackage{amssymb}
\usepackage{multirow}
\usepackage{algorithm}  
\usepackage{algorithmic}
\usepackage{diagbox}
\usepackage{makecell}
\usepackage{booktabs}
\usepackage{xcolor}
\usepackage{colortbl}

\usepackage[english]{babel}  
\modulolinenumbers[50]
\linenumbers

\usepackage[superscript]{cite}
\usepackage{xcolor}
\usepackage{titlesec}
\titleformat*{\section}{\large\bfseries\sffamily}
\titleformat*{\subsection}{\normalsize\bfseries\sffamily}
\titleformat*{\subsubsection}{\small\bfseries\sffamily}

\renewenvironment{abstract}{\global\setbox\absbox=\vbox\bgroup
  \hsize=\textwidth%
  \noindent\unskip\textbf{\large Abstract}
  \par\medskip\noindent\unskip\ignorespaces}{\egroup}
\makeatletter
\renewcommand\@biblabel[1]{#1}
\makeatother
\journal{Lancet}

\begin{document}
\begin{frontmatter}
\title{Efficient endometrial carcinoma screening via cross-modal synthesis and gradient distillation}

\author[addr1,addr2]{Dongjing Shan} 
\author[addr1]{Yamei Luo}
\author[addr3]{Jiqing Xuan}
\author[addr1]{Lu Huang}
\author[addr1]{Jin Li}
\author[addr4]{Mengchu Yang}
\author[addr5]{Zeyu Chen}
\author[addr6]{Fajin Lv}
\author[addr2,addr8,addr9]{Yong Tang\corref{corrauth}}
\cortext[corrauth]{Corresponding authors}
\ead{tangyong@uestc.edu.cn}
\author[addr5,addr7,addr10,addr11]{Chunxiang Zhang\corref{corrauth}}
\ead{zhangchx999@163.com}

\address[addr1]{ School of Medical Information and Engineering, Southwest Medical University, Luzhou 646000, Sichuan, China}
\address[addr2]{ School of Computer Science and Engineering, University of Electronic Science and Technology of China, Chengdu 611731, China}
\address[addr3]{ Department of Ultrasound, Affiliated Hospital of Southwest Medical University, Luzhou 646000, Sichuan, China}
\address[addr4]{ Department of Functional Examination Unit, Zibo Hospital of Traditional Chinese Medicine, Zibo 255000, China}
\address[addr5]{ Key Laboratory of Medical Electrophysiology, Ministry of Education\& Medical Electrophysiological Key Laboratory of Sichuan Province, Institute of Cardiovascular Research, Southwest Medical University}
\address[addr6]{ Department of Radiology, the First Affiliated Hospital of Chongqing Medical University, Chongqing 400016, China}
\address[addr7]{ Department of Cardiology, Affiliated Hospital of Southwest Medical University, Luzhou 646000, Sichuan, China}
\address[addr8]{ International Research Center for Complexity Sciences, Hangzhou International Innovation Institute, Beihang University, Hangzhou 311115, China}
\address[addr9]{ Institute of Intelligent Chinese Medicine, Chongqing University of  Chinese Medicine, Chongqing 402760, China}
\address[addr10]{ Basic Medicine Research Innovation Center for Cardiometabolic Diseases, Ministry of Education, Southwest Medical University}
\address[addr11]{ Nucleic Acid Medicine of Luzhou Key Laboratory, Southwest Medical University, Luzhou 646000, Sichuan, China}

\begin{abstract}
  Early detection of myometrial invasion is critical for the staging and life-saving management of endometrial carcinoma (EC), a prevalent global malignancy. Transvaginal ultrasound serves as the primary, accessible screening modality in resource-constrained primary care settings; however, its diagnostic reliability is severely hindered by low tissue contrast, high operator dependence, and a pronounced scarcity of positive pathological samples. Existing artificial intelligence solutions struggle to overcome this severe class imbalance and the subtle imaging features of invasion, particularly under the strict computational limits of primary care clinics. Here we present an automated, highly efficient two-stage deep learning framework that resolves both data and computational bottlenecks in EC screening. To mitigate pathological data scarcity, we develop a structure-guided cross-modal generation network that synthesizes diverse, high-fidelity ultrasound images from unpaired magnetic resonance imaging (MRI) data, strictly preserving clinically essential anatomical junctions. Furthermore, we introduce a lightweight screening network utilizing gradient distillation, which transfers discriminative knowledge from a high-capacity teacher model to dynamically guide sparse attention towards task-critical regions. Evaluated on a large, multicenter cohort of 7,951 participants, our model achieves a sensitivity of 99.5\%, a specificity of 97.2\%, and an area under the curve of 0.987 at a minimal computational cost (0.289 GFLOPs), substantially outperforming the average diagnostic accuracy of expert sonographers. Our approach demonstrates that combining cross-modal synthetic augmentation with knowledge-driven efficient modeling can democratize expert-level, real-time cancer screening for resource-constrained primary care settings.
\end{abstract}

\end{frontmatter}

\begin{multicols}{2}
Endometrial carcinoma (EC) ranks as the sixth most common malignancy among women globally, with a steadily increasing incidence~\cite{crosbie2022endometrial,bray2024global}. Characterized by its potential for cure in early stages, early screening is critical for improving patient outcomes. Clinical data indicate a 91\%-95\% five-year survival rate for stage I (FIGO) EC patients after standard treatment, often preserving fertility or enabling minimally invasive procedures. In contrast, advanced-stage (FIGO III-IV) patients experience a sharp decline in survival (20\%-50\%) and require radical surgery combined with radiotherapy/chemotherapy, significantly compromising quality of life~\cite{bray2024global,berek2023figo}. Thus, timely detection and accurate assessment of disease progression through early screening are paramount for reducing EC mortality and optimizing therapeutic strategies~\cite{koskas2025cancer}. Current clinical assessment of invasion depth primarily relies on ultrasound. However, the operator-dependent nature and technical constraints of ultrasound, such as low image contrast, textural heterogeneity, artifacts, insufficient image resolution, and difficulty in visualizing micro-infiltrates, limit its capability to reliably distinguish between invasive and non-invasive lesions~\cite{green2018endometrial}, and thus create a pressing need for high-precision, automated image classification models to aid clinical decision-making.


Nevertheless, the development of such assisted diagnostic tools is hindered by two major challenges: \textbf{severe class imbalance and the inherent difficulty of the classification task}. Regarding the first challenge, clinical data exhibits a pronounced long-tail distribution~\cite{balch2024risk,ktena2024generative}: over 90\% of ultrasound samples are from normal endometrium or benign hyperplasia, while cases related to endometrial carcinoma are scarce. Among the carcinoma samples, those with superficial myometrial invasion constitute only 3\%-5\% of the total screening dataset, and samples with deep invasion are even rarer, accounting for less than 1\%. This is because deep invasion often corresponds to mid-advanced stage disease, which has a lower detection rate in early screening. Consequently, models are easily dominated by the overwhelming number of \enquote{normal/superficial invasion} samples, leading to insufficient learning for the critical minority class of \enquote{deep invasion}. This results in a model with high recall but low precision, failing to effectively identify true severe cases and ultimately contradicting the fundamental goal of screening.

The second challenge, detecting myometrial invasion, is compounded by intrinsic limitations of ultrasound imaging. The low resolution typical of primary care devices, along with artifacts from bowel gas or irregular endometrial thickening, often obscures the endometriumšCmyometrium interface~\cite{yang2025saliency}. The lack of definitive, quantifiable sonographic markers further complicates delineation, forcing diagnosis to rely on subtle, subjective features such as textural, echogenic, or morphological changes at the tumor boundary. While an intact junction appears as a continuous hypoechoic band, invasion is suggested by disruption or fusion. Discriminating these nuanced patterns demands high-capacity feature extraction, an obstacle for lightweight models with limited representational power, increasing the risk of misclassification.

Addressing the core clinical needs for primary care EC ultrasound screening, this paper establishes a Lightweight Screening Network based on ultrasound modality for the precise binary classification of myometrial invasion (presence vs. absence). Our approach specifically targets the aforementioned challenges: To tackle class imbalance, we propose a \textbf{Structure-Guided Cycle-Consistent Adversarial Network} (SG-CycleGAN) to synthesize high-quality, diverse MRI-to-ultrasound images. This approach is clinically grounded, as patients with suspected ultrasound findings often undergo further MRI examination, making MRI-derived synthetic ultrasound a viable strategy to compensate for the scarcity of pathological ultrasound samples. To overcome the inherent challenges in distinguishing subtle invasion patterns, we introduce \textbf{Gradient Distillation} as a core technical solution that systematically enhances the lightweight network's discriminative capacity through dual mechanisms of knowledge transfer and attention sparsification, thereby achieving both improved diagnostic accuracy and computational efficiency.

\section*{{Results}}
\subsection*{Data}
This multicenter retrospective study, conducted across The First Affiliated Hospital of Chongqing Medical University, The Affiliated Hospital of Southwest Medical University, Zibo Hospital of Traditional Chinese Medicine, Zibo Maternal and Child Health Hospital, and Jincheng People's Hospital between March 2018 and September 2025, analyzed transvaginal sonography (TVS) data comprising 3,354 images from 651 patients with histologically confirmed Endometrial Carcinoma (EC), alongside 33,189 images from 7,300 women with normal uteri. The overall cohort consisted of 7,951 participants (651 EC patients and 7,300 controls). All TVS images from EC patients were obtained within 7 days before surgery. Inclusion criteria for EC patients comprised: 1) postoperative pathological confirmation of endometrial carcinoma and 2) completion of a preoperative TVS examination within one week prior to surgery. Exclusion criteria were: 1) coexistence of other malignancies, 2) obstruction of lesions by Doppler flow signals, and 3) incomplete image series.

To ensure data distribution balance across centers and avoid overfitting to specific institutional characteristics, the ultrasound dataset was randomly partitioned at the patient level into training, validation, and test sets following an 8:1:1 ratio (no center-specific partitioning was adopted). All participating centers followed a unified TVS examination protocol (in accordance with the International Society of Ultrasound in Obstetrics and Gynecology [ISUOG] practice guidelines) and implemented a standardized image cropping method to remove institutional logos, text, and other irrelevant information, thereby minimizing inter-institutional heterogeneity caused by non-diagnostic institutional-specific markers. This partitioning strategy ensured that each dataset (training, validation, test) contained representative samples from all five centers, reflecting the diverse clinical scenarios of EC diagnosis in routine practice. This resulted in a training set containing 2,698 EC images and 23,219 normal images. To ensure balanced evaluation, the validation and test sets were constructed to contain 320 and 336 EC images, respectively, with an equal number of randomly selected normal images (320 and 336, respectively) included in each dataset.

In parallel, this study retrospectively collected MRI data from 498 surgically confirmed endometrial carcinoma patients. All participants underwent multi-sequence MRI scanning, which included axial T2-weighted imaging, axial T1-weighted imaging (both non-contrast and fat-suppressed post-contrast sequences), axial diffusion-weighted imaging (DWI) with corresponding apparent diffusion coefficient (ADC) maps, and axial LAVA dynamic contrast-enhanced sequences.
The institutional review board approved this study, and informed consent was waived due to its retrospective design.
After image preprocessing, a total of 34,214 images were extracted from different scanning sequences and slices. To ensure rigorous model evaluation, all data were randomly partitioned at the patient level. Specifically, the 498 patients were randomly allocated into training and test sets in a 2:1 ratio. A key principle of this partitioning was to ensure the complete independence of patient IDs in the test set, meaning that all images corresponding to these patients were excluded from any stage of model training.
\begin{table*}[t]
\centering
\renewcommand{\arraystretch}{1.5}
\begin{threeparttable}
\caption{Comparison of Image Quality Metrics.}
\label{tab:image_quality_metrics}
\begin{tabular}{|c|c|c|c|c|c|}
\hline
\diagbox{Metric}{Method} & {CycleGAN~\cite{zhu2017unpaired}} & {UNIT~\cite{liu2017unsupervised}} & {MUNIT~\cite{huang2018multimodal}} & {DCLGAN~\cite{han2021dual}} & {SG-CycleGAN} \\ \hline
{Params (M)} & 220 & 217 & 196 & 227 & 213 \\ \hline
{FID $\downarrow$} & 133.2858 & 219.1379 & 165.5644 & 122.9870 & \textbf{73.2540} \\ \hline
\multirow{2}{*}{{KID $\downarrow$}} & 0.1128 & 0.2588 & 0.1771 & 0.0947 & \textbf{0.0636} \\
\cline{2-6}
& $\pm$0.0026 & $\pm$0.0031 & $\pm$0.0042 & $\pm$0.0020 & $\pm$0.0015 \\ \hline
\end{tabular}
\begin{tablenotes}
\footnotesize
\item \textit{Note: SD (Standard Deviation).}  
\end{tablenotes}
\end{threeparttable}
\end{table*}

\subsection*{Superiority of SG-CycleGAN over Baseline CycleGANs}
Building upon the dataset described in the previous section, we designed a comparative experiment to rigorously evaluate the performance of our proposed SG-CycleGAN. While generative adversarial networks~\cite{waikel2024recognition,ali2025generative} are inherently bidirectional, our specific application scenario required only unidirectional translation from MRI to ultrasound. Therefore, for the adversarial training of all compared models, we utilized the entire set of 2,634 ultrasound images from histologically confirmed endometrial carcinoma cases and 22,809 MRI images, corresponding to two-thirds of the total MRI dataset. The remaining one-third of the MRI images were reserved for the test set to generate synthetic ultrasound images for evaluation. The primary baseline models for comparison included UNIT~\cite{liu2017unsupervised} (Unsupervised Image-to-Image Translation), MUNIT~\cite{huang2018multimodal} (Multimodal UNsupervised Image-to-Image Translation) , and DCLGAN~\cite{han2021dual} (Dual Contrastive Learning for Unpaired Image-to-Image Translation). However, other more recent models, such as those for unpaired CT$\leftrightarrow$MR image synthesis (CycleSGAN~\cite{wang2024cyclesgan}, DC-CycleGAN~\cite{wang2023dc}) or multi-domain style translation (StarGAN v2~\cite{choi2020stargan}), along with the recently proposed Onto-CGAN~\cite{sun2025generating} (a model that generates synthetic EHR data for unseen diseases by integrating domain knowledge), were not suitable for ultrasound-to-MRI image translation. To comprehensively assess the quality of the synthesized images and their utility for the downstream task, we employed a multifaceted evaluation protocol focusing on two key aspects:

1) \textbf{Image Quality Metrics}: We used the Fr\'{e}chet Inception Distance~\cite{chong2020effectively} (FID) and the Kernel Inception Distance~\cite{jimenez2024gan} (KID) to quantitatively measure the similarity between the distribution of synthetic ultrasound images and the distribution of real ultrasound images. Lower FID and KID scores indicate that the generated images are more photorealistic and statistically closer to real ultrasound scans, which is crucial for ensuring the perceptual validity of the augmented data.

To ensure a fair performance comparison, the generators of all compared models were balanced in architectural design to maintain a comparable parameter scale, as shown in Table \ref{tab:image_quality_metrics}. Our proposed SG-CycleGAN, which incorporates a shared modality-agnostic feature extractor (MAFE), contains 213M parameters in its unidirectional generator. The parameter count of SG-CycleGAN was kept close to those of the other models by using shallower private branches with a reduced number of residual blocks (set to 7), whereas the standard CycleGAN was expanded to 9 residual blocks to increase its capacity. This deliberate balancing ensures that the models are compared under a similar parameter budget, so that the performance differences reported in Table \ref{tab:image_quality_metrics} primarily stem from the proposed structure guidance and gradient distillation mechanisms, rather than from disparities in model capacity.

As presented in Table \ref{tab:image_quality_metrics}, the quantitative results demonstrate the superior performance of our SG-CycleGAN in generating realistic and structurally faithful synthetic ultrasound images. SG-CycleGAN achieves the lowest Fr\'{e}chet Inception Distance (FID) of 73.25 and the lowest Kernel Inception Distance (KID) of 0.0636$\pm$0.0015, indicating that its synthetic image distribution is statistically closest to that of real ultrasound scans. In comparison, the baseline CycleGAN yields an FID of 133.29 and a KID of 0.1128, while other contemporary models (UNIT, MUNIT, and DCLGAN) exhibit higher or moderately competitive scores. Notably, DCLGAN, which employs dual contrastive learning, attains an FID of 122.99 and a KID of 0.0947, yet remains outperformed by our method.
\begin{figure*}[ht]
    \centering
    \begin{subfigure}[b]{0.155\textwidth}
        \centering
        \includegraphics[width=\textwidth]{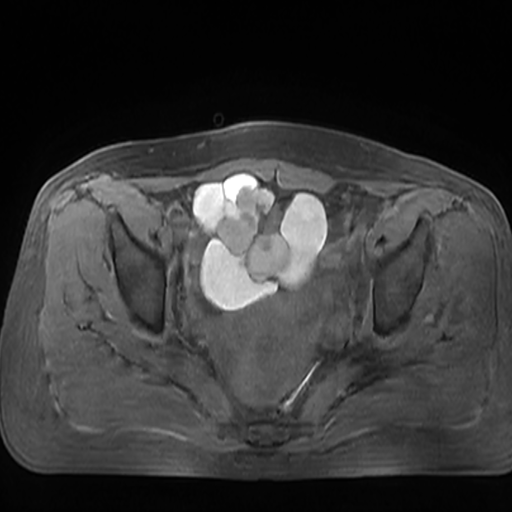}
        \label{fig:mri_original_1}
    \end{subfigure}
    \hfill
    \begin{subfigure}[b]{0.155\textwidth}
        \centering
        \includegraphics[width=\textwidth]{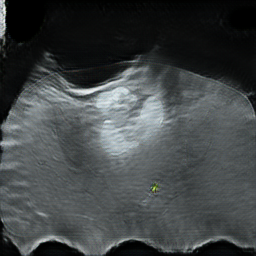}
        \label{fig:cyclegan_1}
    \end{subfigure}
    \hfill
    \begin{subfigure}[b]{0.155\textwidth}
        \centering
        \includegraphics[width=\textwidth]{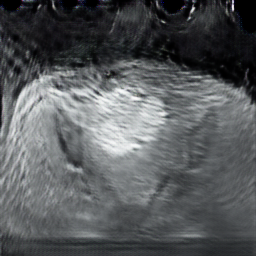}
        \label{fig:unit_1}
    \end{subfigure}
    \hfill
    \begin{subfigure}[b]{0.155\textwidth}
        \centering
        \includegraphics[width=\textwidth]{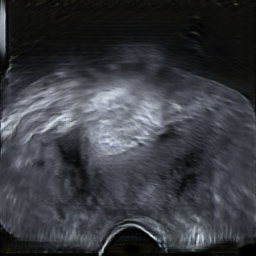}
        \label{fig:munit_1}
    \end{subfigure}
    \hfill
    \begin{subfigure}[b]{0.155\textwidth}
        \centering
        \includegraphics[width=\textwidth]{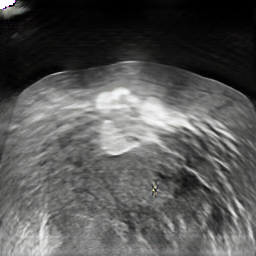}
        \label{fig:dclgan_1}
    \end{subfigure}
    \hfill
    \begin{subfigure}[b]{0.155\textwidth}
        \centering
        \includegraphics[width=\textwidth]{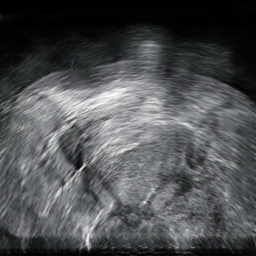}
        \label{fig:sg_cyclegan_1}
    \end{subfigure}

    \vspace{-5pt}

    \begin{subfigure}[b]{0.155\textwidth}
        \centering
        \includegraphics[width=\textwidth]{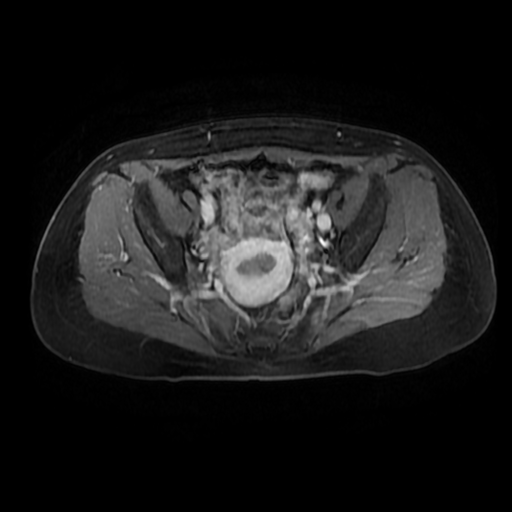}
        \caption{MRI}
        \label{fig:mri_original_2}
    \end{subfigure}
    \hfill
    \begin{subfigure}[b]{0.155\textwidth}
        \centering
        \includegraphics[width=\textwidth]{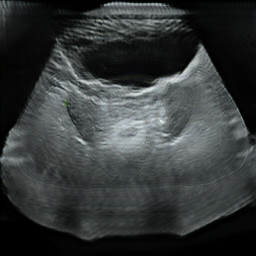}
        \caption{CycleGAN}
        \label{fig:cyclegan_2}
    \end{subfigure}
    \hfill
    \begin{subfigure}[b]{0.155\textwidth}
        \centering
        \includegraphics[width=\textwidth]{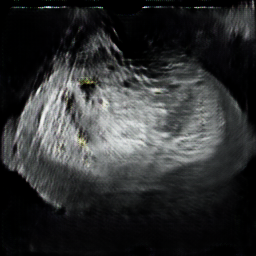}
        \caption{UNIT}
        \label{fig:unit_2}
    \end{subfigure}
    \hfill
    \begin{subfigure}[b]{0.155\textwidth}
        \centering
        \includegraphics[width=\textwidth]{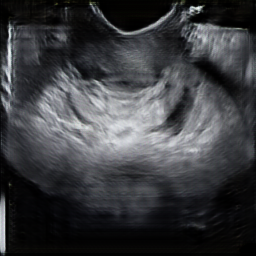}
        \caption{MUNIT}
        \label{fig:munit_2}
    \end{subfigure}
    \hfill
    \begin{subfigure}[b]{0.155\textwidth}
        \centering
        \includegraphics[width=\textwidth]{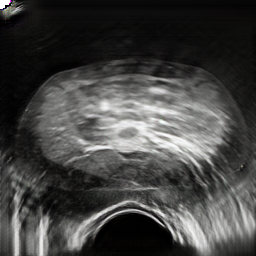}
        \caption{DCLGAN}
        \label{fig:dclgan_2}
    \end{subfigure}
    \hfill
    \begin{subfigure}[b]{0.155\textwidth}
        \centering
        \includegraphics[width=\textwidth]{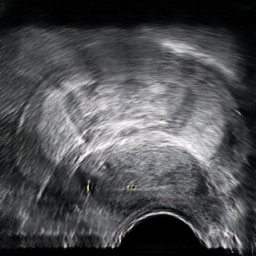}
        \caption{SG-CycleGAN}
        \label{fig:sg_cyclegan_2}
    \end{subfigure}

    \caption{Visual comparison of MRI-to-ultrasound image synthesis results from different models. (a) Original axial MR images of the uterus. (b-f) Corresponding synthetic ultrasound images generated by five comparative models, respectively.}
    \label{fig:image_synthesis_comparison}
\end{figure*}

These improvements can be attributed to the structure-preserving mechanisms embedded in SG-CycleGAN, namely the modality-agnostic feature extractor (MAFE) and the feature consistency loss, which explicitly maintain anatomical integrity during cross-modal translation. Under a comparable parameter budget, the consistent gains in both FID and KID metrics confirm that the enhanced synthesis quality stems from the proposed architectural innovations rather than increased model capacity, thereby validating the effectiveness of our approach for generating diagnostically relevant ultrasound data. Examples of cross-modal image translation are presented in Figure \ref{fig:image_synthesis_comparison}, where panel (a) shows the original MRI, and the five columns on the right display the corresponding ultrasound images generated by the five compared models respectively. While other methods (CycleGAN, MUNIT, DCLGAN) often produce smoothed or artifact-prone ultrasound-like outputs, and UNIT notably fails to generate plausible ultrasound appearance, the SG-CycleGAN synthesizes images with superior structural fidelity and more realistic textural details. This visual assessment is consistent with the quantitative superiority reported in Table \ref{tab:image_quality_metrics}.

\begin{figure*}[ht]
    \centering
    \begin{subfigure}[b]{0.155\textwidth}
        \centering
        \includegraphics[width=\textwidth]{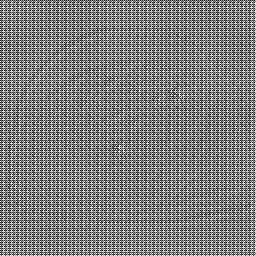}
    \end{subfigure}
    \hfill
    \begin{subfigure}[b]{0.155\textwidth}
        \centering
        \includegraphics[width=\textwidth]{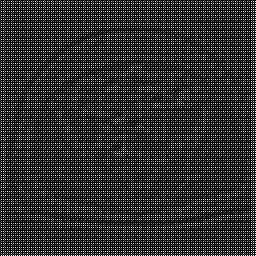}
    \end{subfigure}
    \hfill
    \begin{subfigure}[b]{0.155\textwidth}
        \centering
        \includegraphics[width=\textwidth]{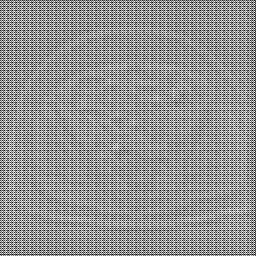}
    \end{subfigure}
    \hfill
    \begin{subfigure}[b]{0.155\textwidth}
        \centering
        \includegraphics[width=\textwidth]{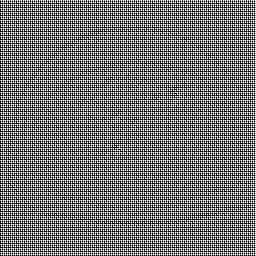}
    \end{subfigure}
    \hfill
    \begin{subfigure}[b]{0.155\textwidth}
        \centering
        \includegraphics[width=\textwidth]{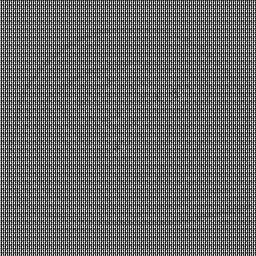}
    \end{subfigure}
    \hfill
    \begin{subfigure}[b]{0.155\textwidth}
        \centering
        \includegraphics[width=\textwidth]{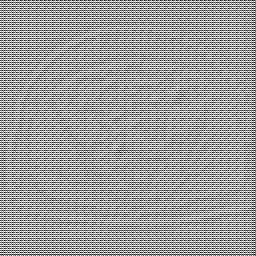}
    \end{subfigure}
    \caption{Visualization of selected feature maps from the modality-agnostic feature extractor (MAFE).}
    \label{fig:MAFE_image}
\end{figure*}

To visually demonstrate the structural-preserving capability of our SG-CycleGAN, Figure \ref{fig:MAFE_image} presents the modality-agnostic feature maps extracted by the MAFE module. By discarding modality-specific textures and retaining only shared structural information, the MAFE ensures that the synthesized ultrasound images preserve the critical anatomical layout necessary for reliable invasion assessment, thereby underpinning the improved synthesis quality shown in Figure \ref{fig:image_synthesis_comparison} and the downstream classification performance reported in Table \ref{tab:downstream_classification}.

2) \textbf{Downstream Task Performance}: To assess the clinical utility of the synthetic ultrasound data beyond perceptual quality, we conducted a downstream binary classification task for myometrial invasion. A lightweight MobileNet-V2~\cite{sandler2018mobilenetv2} classifier (comprising 730.5K parameters) was trained from scratch using a hybrid dataset: positive samples were comprised of 10,000 real ultrasound images randomly selected from the constructed training set, while negative samples consisted of synthetic ultrasound images generated from MRI data. This design evaluates the model's ability to identify disease samples after being trained on a hybrid dataset comprising both real and synthetic ultrasound data. For model evaluation, we utilized the constructed validation and test sets, both consisting exclusively of held-out real ultrasound images that were not used during training. The model parameters yielding the highest sensitivity on the validation set were selected and saved. These optimized parameters were then applied to the independent test set for the final performance assessment.
To ensure reproducibility, all experiments were independently repeated three times using distinct random seeds (42, 142, 242), following established practices in machine learning research~\cite{henderson2018deep,recht2019imagenet}.
The mean and standard deviation of each metric from these three runs are presented in Table \ref{tab:downstream_classification}.
\begin{table*}[ht]
\centering
\caption{Comparison of Downstream Classification Task Performance (MobileNet-V2 on Synthetic Data).}
{
\label{tab:downstream_classification}
\renewcommand{\arraystretch}{1.5}
\begin{tabular}{|l|c|c|c|c|c|}
\hline
{Metric} & {CycleGAN~\cite{zhu2017unpaired}} & {UNIT~\cite{liu2017unsupervised}} & {MUNIT~\cite{huang2018multimodal}} & {DCLGAN~\cite{han2021dual}} & {SG-CycleGAN} \\ \hline
{Accuracy} & 0.7431 \textcolor{darkgray}{$\pm$ 0.0322} & 0.7465 \textcolor{darkgray}{$\pm$ 0.0049} & 0.5833 \textcolor{darkgray}{$\pm$ 0.0279} & 0.7240 \textcolor{darkgray}{$\pm$ 0.0533} & \textbf{0.7691} \textcolor{darkgray}{$\pm$ 0.0302} \\ \hline
{Sensitivity} & 0.7188 \textcolor{darkgray}{$\pm$ 0.0450} & 0.7882 \textcolor{darkgray}{$\pm$ 0.0547} & 0.3958 \textcolor{darkgray}{$\pm$ 0.1179} & 0.6424 \textcolor{darkgray}{$\pm$ 0.1702} & \textbf{0.8160} \textcolor{darkgray}{$\pm$ 0.0214} \\ \hline
{Specificity} & 0.7674 \textcolor{darkgray}{$\pm$ 0.0260} & 0.7049 \textcolor{darkgray}{$\pm$ 0.0491} & 0.7708 \textcolor{darkgray}{$\pm$ 0.0696} & \textbf{0.8056} \textcolor{darkgray}{$\pm$ 0.0650} & 0.7222 \textcolor{darkgray}{$\pm$ 0.0772} \\ \hline
{Precision} & 0.7551 \textcolor{darkgray}{$\pm$ 0.0300} & 0.7293 \textcolor{darkgray}{$\pm$ 0.0184} & 0.6340 \textcolor{darkgray}{$\pm$ 0.0272} & \textbf{0.7707} \textcolor{darkgray}{$\pm$ 0.0131} & 0.7500 \textcolor{darkgray}{$\pm$ 0.0464} \\ \hline
{F1-score} & 0.7362 \textcolor{darkgray}{$\pm$ 0.0364} & 0.7560 \textcolor{darkgray}{$\pm$ 0.0149} & 0.4777 \textcolor{darkgray}{$\pm$ 0.0843} & 0.6870 \textcolor{darkgray}{$\pm$ 0.0985} & \textbf{0.7802} \textcolor{darkgray}{$\pm$ 0.0192} \\ \hline
{ROC AUC} & 0.8177 \textcolor{darkgray}{$\pm$ 0.0178} & 0.8205 \textcolor{darkgray}{$\pm$ 0.0360} & 0.6848 \textcolor{darkgray}{$\pm$ 0.0337} & 0.8252 \textcolor{darkgray}{$\pm$ 0.0558} & \textbf{0.8416} \textcolor{darkgray}{$\pm$ 0.0188} \\ \hline
\end{tabular}
}
\end{table*}

A comparative analysis of Table \ref{tab:downstream_classification} delineates distinct performance profiles among the benchmark models. MUNIT and DCLGAN exhibit notably high specificity (0.7708$\pm$0.0696, and 0.8056$\pm$0.0650, respectively) but substantially low sensitivity (0.3958$\pm$0.1179, and 0.6424$\pm$0.1702). This pattern suggests a pronounced bias toward negative predictions, indicative of synthetic images that lack discriminative features for myometrial invasion, thereby leading the classifier to adopt an overly conservative diagnostic strategy. CycleGAN and UNIT demonstrate relatively higher sensitivity (0.7188$\pm$0.0450, and 0.7882$\pm$0.0547) compared to the aforementioned models, while maintaining moderate specificity (0.7674$\pm$0.0260, and 0.7049$\pm$0.0491), reflecting an alternative trade-off that reduces false negatives at the expense of increased false positives.	

Notably, the proposed SG-CycleGAN achieves the most balanced and clinically favorable performance, attaining the highest sensitivity (0.8299$\pm$0.0214) along with a competitive specificity (0.7083$\pm$0.0390). It also yields the best accuracy (0.7691$\pm$0.0098) and F1-score (0.7824$\pm$0.0045) among all compared methods. These results underscore that the structure-preserving synthesis facilitated by SG-CycleGAN enriches the pathological relevance of generated features, enabling more reliable discrimination of invasion without substantial compromise in either sensitivity or specificity. Consequently, the proposed framework validates its efficacy as a robust strategy for augmenting scarce and imbalanced ultrasound datasets in endometrial carcinoma screening.
\begin{figure*}[ht]
    \centering
    \begin{subfigure}[b]{0.1961\textwidth}
        \centering
        \includegraphics[width=\textwidth]{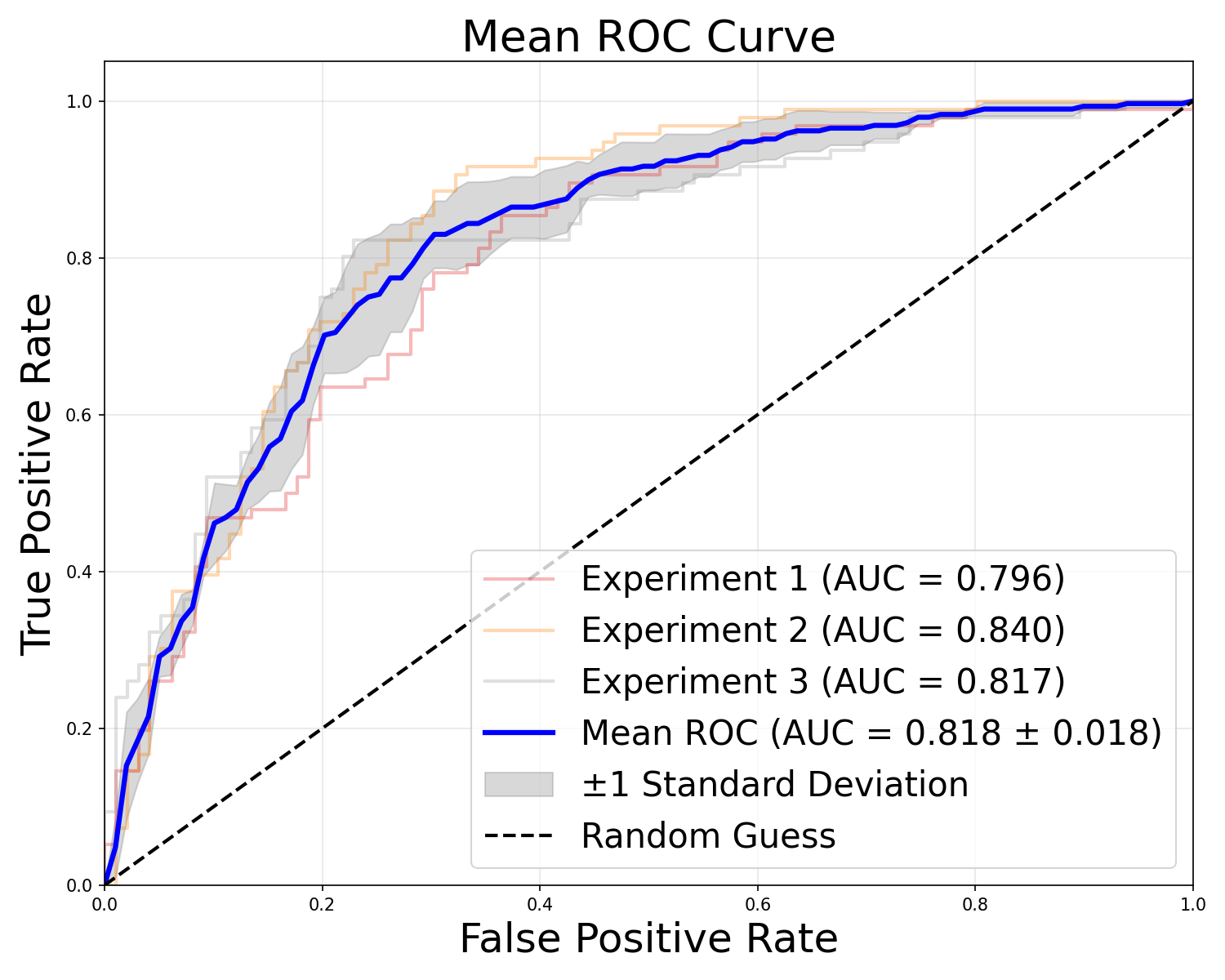}
        \caption{CycleGAN}
    \end{subfigure}
    \hfill
    \begin{subfigure}[b]{0.1961\textwidth}
        \centering
        \includegraphics[width=\textwidth]{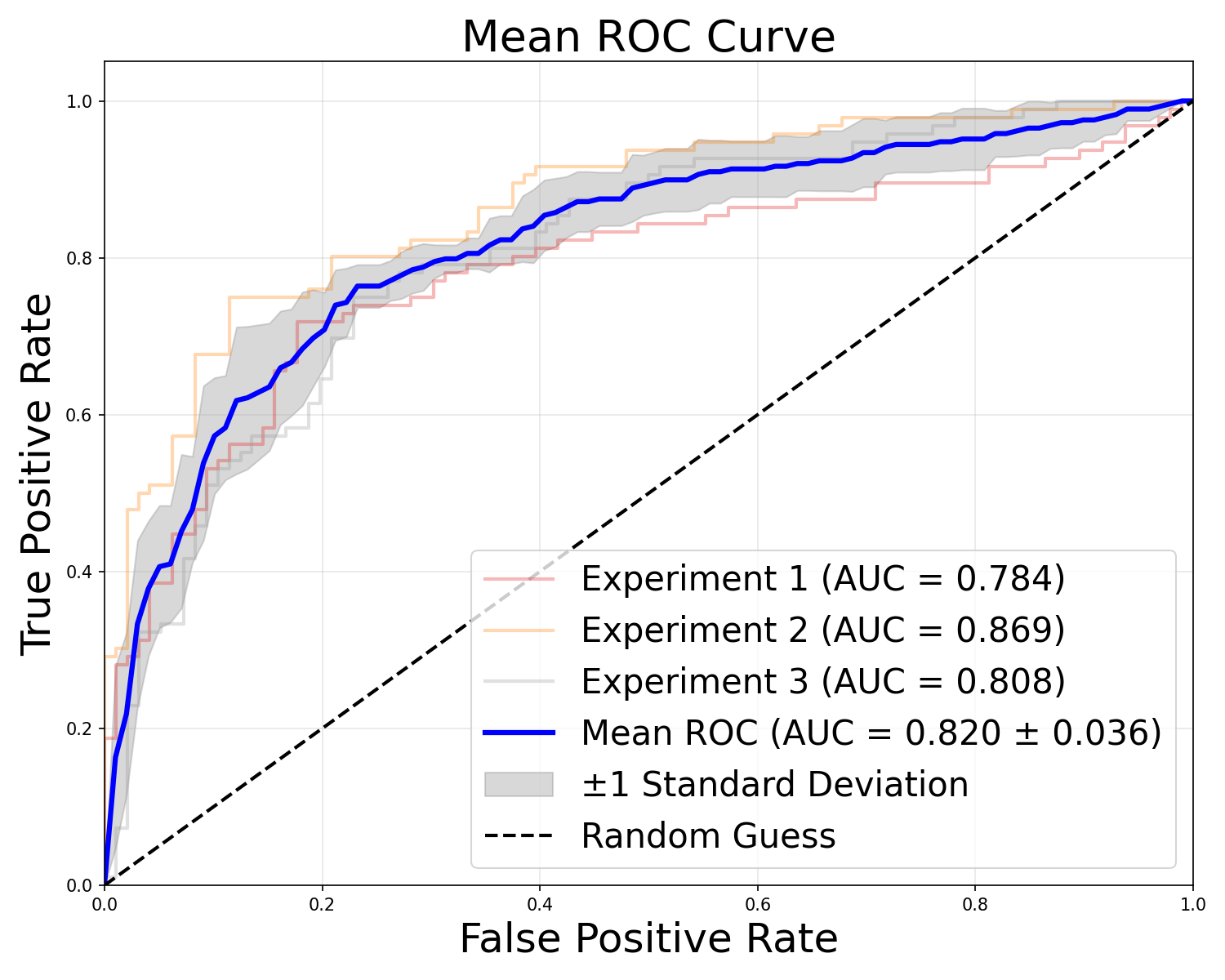}
        \caption{UNIT}
    \end{subfigure}
    \hfill
    \begin{subfigure}[b]{0.1961\textwidth}
        \centering
        \includegraphics[width=\textwidth]{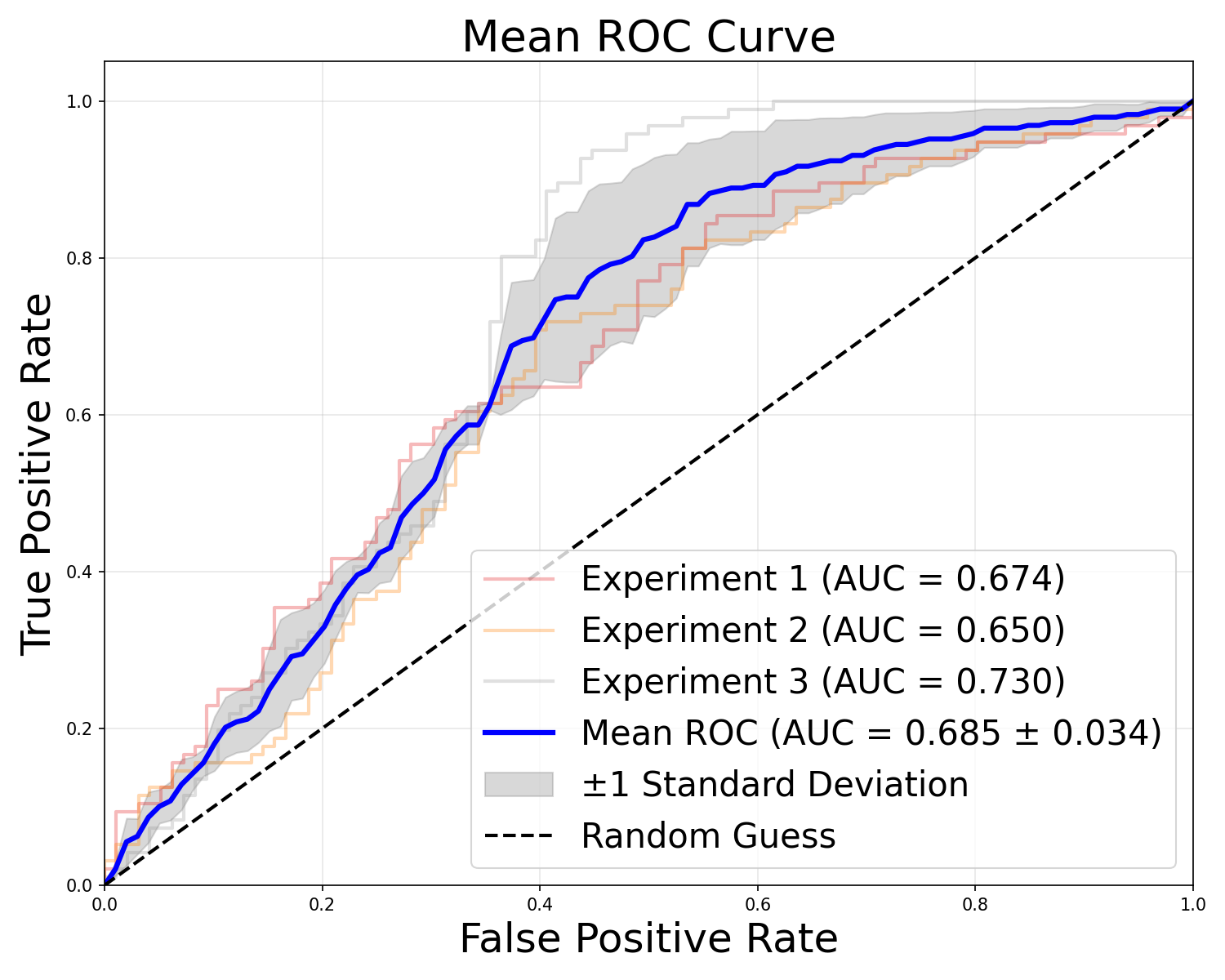}
        \caption{MUNIT}
    \end{subfigure}
    \hfill
    \begin{subfigure}[b]{0.1961\textwidth}
        \centering
        \includegraphics[width=\textwidth]{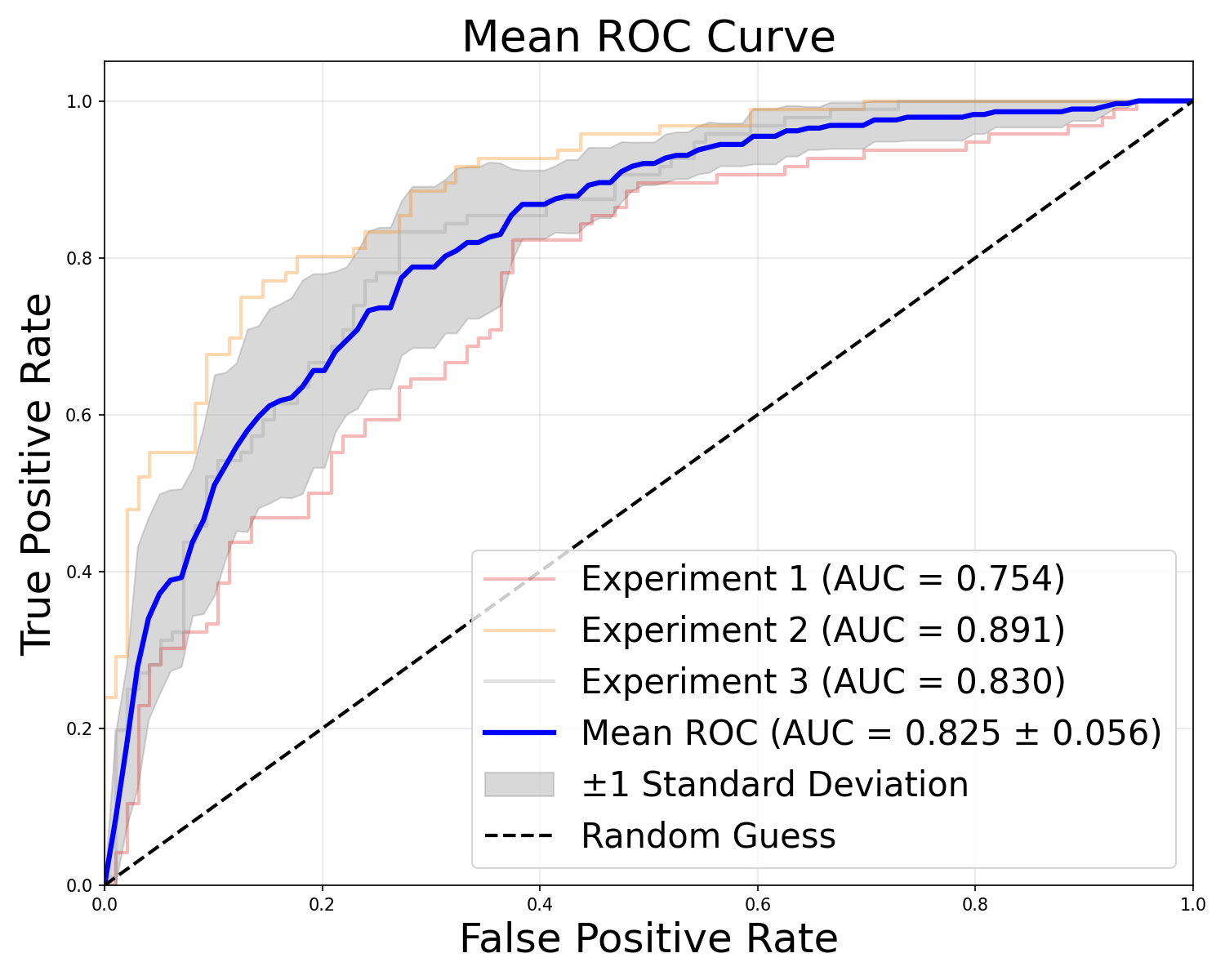}
        \caption{DCLGAN}
    \end{subfigure}
    \hfill
    \begin{subfigure}[b]{0.1961\textwidth}
        \centering
        \includegraphics[width=\textwidth]{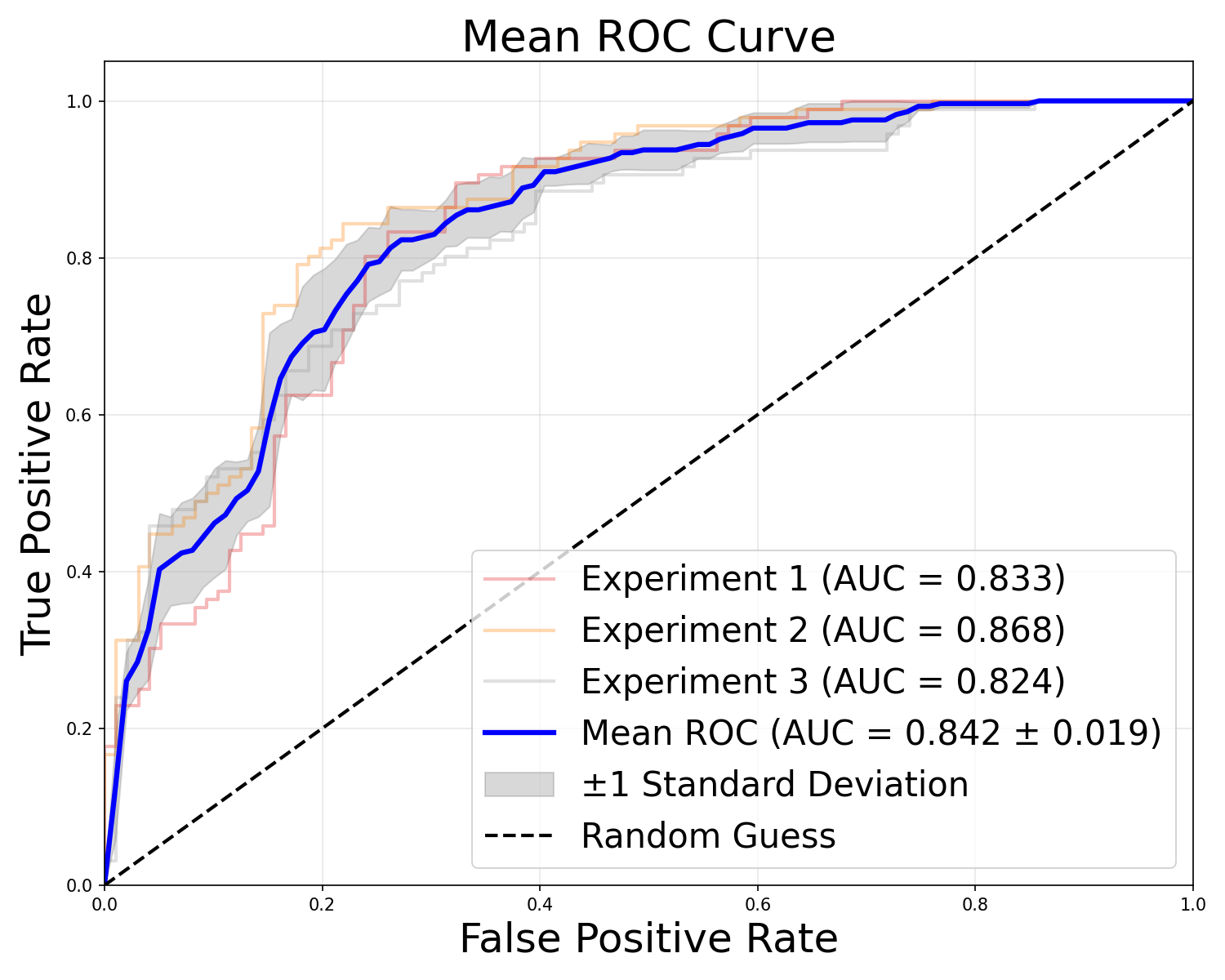}
        \caption{SG-CycleGAN}
    \end{subfigure}

    \caption{Receiver operating characteristic (ROC) curves of the downstream invasion classifier trained on synthetic data. The lightweight MobileNet-V2 classifier was trained using synthetic ultrasound images generated by (a) CycleGAN, (b) UNIT, (c) MUNIT, (d) DCLGAN, and (e) the proposed SG-CycleGAN.}
    \label{fig:roc_curves}
\end{figure*}

Figure~\ref{fig:roc_curves} displays the mean receiver operating characteristic (ROC) curves, each accompanied by a shaded region representing the $\pm$1 standard deviation interval, for the downstream MobileNet-V2 classifier trained on synthetic data from different generative models. These curves and their variability estimates are derived from three completely independent runs of the full training-validation-testing pipeline, ensuring a robust evaluation of each model's discriminative capability. The proposed SG-CycleGAN consistently yields the superior mean ROC curve, attaining the highest mean area under the curve (AUC) of 0.8402. This result, coupled with its narrow standard deviation band, indicates that our method generates the most stable and pathologically discriminative synthetic features, enabling the classifier to achieve reliable and reproducible performance in distinguishing myometrial invasion.

\subsection*{Performance Evaluation of the Lightweight Screening Network}
To comprehensively evaluate the effectiveness of the proposed Lightweight Screening Network (LSNet), we conducted a comparative analysis against several established lightweight architectures, namely MobileNet-V2~\cite{sandler2018mobilenetv2}, EfficientNet~\cite{tan2019efficientnet}, and MobileViT~\cite{mehta2022mobilevit}. All models were trained using a two-phase strategy: pre-training on a mixed dataset followed by fine-tuning on real ultrasound data. In the performance comparison of lightweight models, we kept FLOPs~\cite{chen2023run} (floating-point operations) consistent across all architectures. This setting is more reasonable and equitable than maintaining roughly the same number of parameters, since FLOPs more directly reflect actual computational cost and exhibit a higher correlation with inference speed.

A large-scale hybrid pre-training dataset was constructed by augmenting the available real ultrasound data with synthetic images generated by the proposed SG-CycleGAN. This mixed dataset comprised approximately 13,000 pathological samples (including both real and synthetic images of EC cases) and 23,219 normal ultrasound samples. The training protocol followed a two-phase strategy: first, all models were pre-trained on this hybrid dataset; subsequently, they were fine-tuned using only the held-out real ultrasound data to adapt them to the authentic imaging distribution. All image samples were resized to 320$\times$320 pixels as part of the preprocessing pipeline.

It is important to note that the student model of our proposed LSNet, which includes the integrated gradient simulator, contains 391.8K parameters. In contrast, its corresponding teacher model employs a significantly larger architecture with 6.40M parameters, facilitating effective gradient distillation during the training process. The model selection strategy was optimized for screening performance: model parameters achieving the highest sensitivity on the validation set were saved when sensitivity was below 0.95; when sensitivity exceeded 0.95, parameters yielding the highest specificity were retained instead. This dual-threshold approach ensures both high disease detection (when sensitivity is inadequate) and minimization of false positives (when sensitivity is sufficiently high), thereby optimizing overall screening efficacy and resource utilization in clinical practice.

\begin{table*}[t]
\centering
\caption{Performance Comparison of the Proposed Lightweight Screening Network (LSNet) versus Benchmark Models on Synthetic and Real Ultrasound Data.}
{
\renewcommand{\arraystretch}{1.5}
\label{tab:lsnet_comparison}
\begin{tabular}{|l|c|c|c|c|}
\hline
{Metric} & {MobileNet-V2~\cite{sandler2018mobilenetv2}} & {EfficientNet~\cite{tan2019efficientnet}} & {MobileViT~\cite{mehta2022mobilevit}} & {LSNet (Ours)} \\ \hline
{FLOPs} & 0.338G & 0.513G & 0.562G & \textbf{0.289}G \\ \hline
{Params} & 1.013M & 204.5K & 294.5K & 391.8K \\ \hline
{Accuracy} & 0.9499 \textcolor{darkgray}{$\pm$ 0.0092} & 0.9345 \textcolor{darkgray}{$\pm$ 0.0263} & 0.9688 \textcolor{darkgray}{$\pm$ 0.0080} & \textbf{0.9836} \textcolor{darkgray}{$\pm$ 0.0024} \\ \hline
{Sensitivity} & 0.9573 \textcolor{darkgray}{$\pm$ 0.0148} & 0.9028 \textcolor{darkgray}{$\pm$ 0.0552} & 0.9762 \textcolor{darkgray}{$\pm$ 0.0106} & \textbf{0.9950} \textcolor{darkgray}{$\pm$ 0.0014} \\ \hline
{Specificity} & 0.9425 \textcolor{darkgray}{$\pm$ 0.0140} & 0.9663 \textcolor{darkgray}{$\pm$ 0.0028} & 0.9613 \textcolor{darkgray}{$\pm$ 0.0084} & \textbf{0.9722} \textcolor{darkgray}{$\pm$ 0.0051} \\ \hline
{Precision} & 0.9435 \textcolor{darkgray}{$\pm$ 0.0127} & 0.9640 \textcolor{darkgray}{$\pm$ 0.0012} & 0.9619 \textcolor{darkgray}{$\pm$ 0.0081} & \textbf{0.9729} \textcolor{darkgray}{$\pm$ 0.0048} \\ \hline
{F1-score} & 0.9503 \textcolor{darkgray}{$\pm$ 0.0092} & 0.9315 \textcolor{darkgray}{$\pm$ 0.0295} & 0.9690 \textcolor{darkgray}{$\pm$ 0.0079} & \textbf{0.9838} \textcolor{darkgray}{$\pm$ 0.0024} \\ \hline
{ROC AUC} & 0.9812 \textcolor{darkgray}{$\pm$ 0.0057} & 0.9790 \textcolor{darkgray}{$\pm$ 0.0008} & 0.9811 \textcolor{darkgray}{$\pm$ 0.0023} & \textbf{0.9873} \textcolor{darkgray}{$\pm$ 0.0012} \\ \hline
\end{tabular}
}
\end{table*}

Based on the comparative results presented in Table \ref{tab:lsnet_comparison}, the proposed Lightweight Screening Network (LSNet) demonstrates superior diagnostic performance while maintaining the lowest computational footprint. LSNet achieves the highest accuracy (0.9836$\pm$0.0024), specificity (0.9722$\pm$0.0051), precision (0.9729$\pm$0.0048), F1-score (0.9790$\pm$0.0018), and ROC-AUC (0.9920$\pm$0.0005), coupled with a leading sensitivity (0.9950$\pm$0.0014). Notably, this performance is attained with only 0.289 GFLOPs and 391.8K parameters, which are lower or comparable to the benchmark models (MobileNet-V2, EfficientNet, and MobileViT). This indicates that LSNet successfully balances high accuracy with operational efficiency. The performance gain can be attributed to its integrated gradient distillation and sparse attention mechanism, which effectively transfer discriminative knowledge from a larger teacher model and focus computational resources on pathologically relevant regions. Furthermore, an average inference time of 0.157 seconds per image on an Intel Core i9 CPU (3.5GHz base frequency), compared to 0.262 seconds for MobileViT, underscores the practical viability of LSNet for real-time screening in resource-constrained primary care settings.

\begin{figure*}[ht]
    \centering
    \begin{subfigure}[b]{0.246\textwidth}
        \centering
        \includegraphics[width=\textwidth]{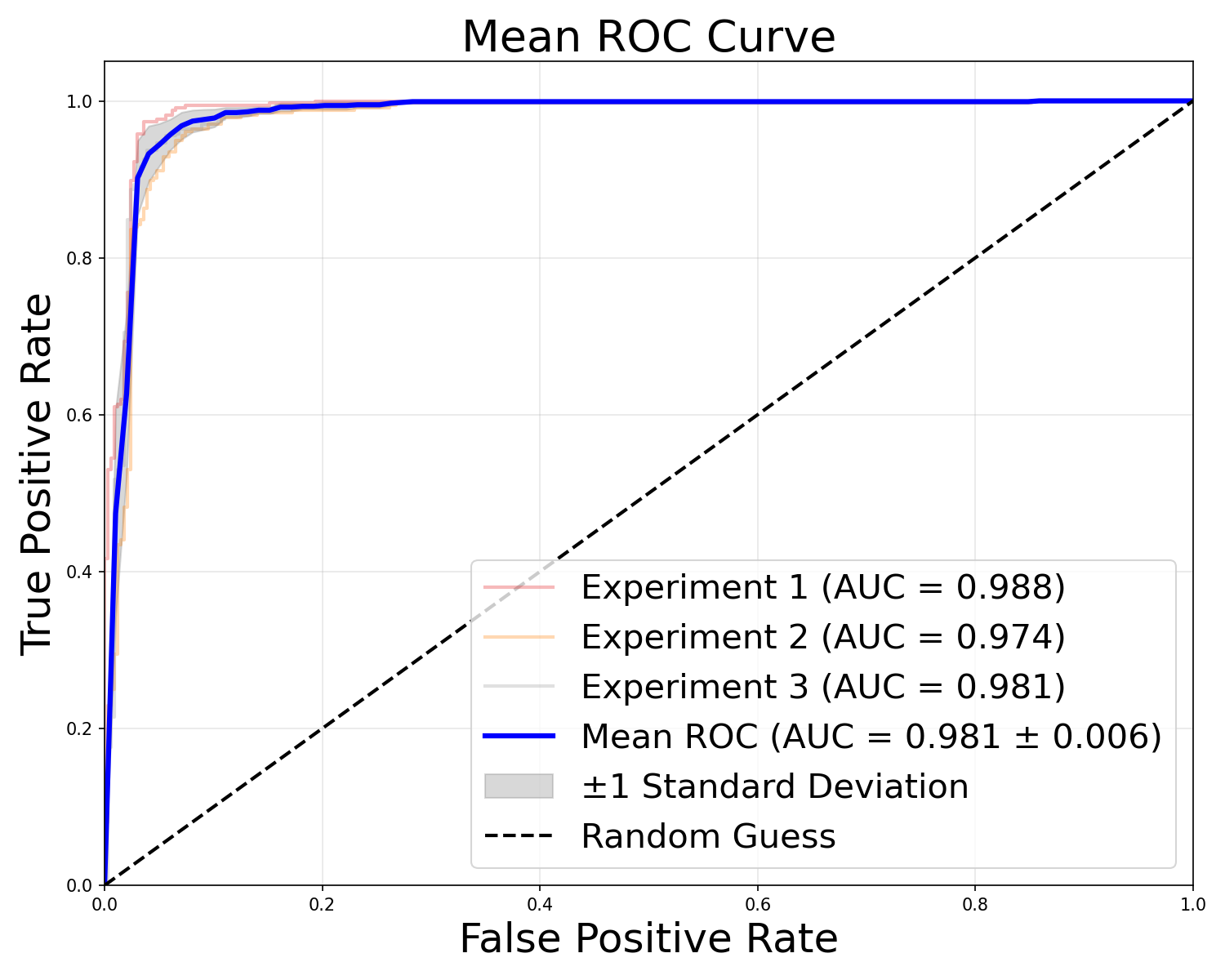}
        \caption{CycleGAN}
    \end{subfigure}
    \hfill
    \begin{subfigure}[b]{0.246\textwidth}
        \centering
        \includegraphics[width=\textwidth]{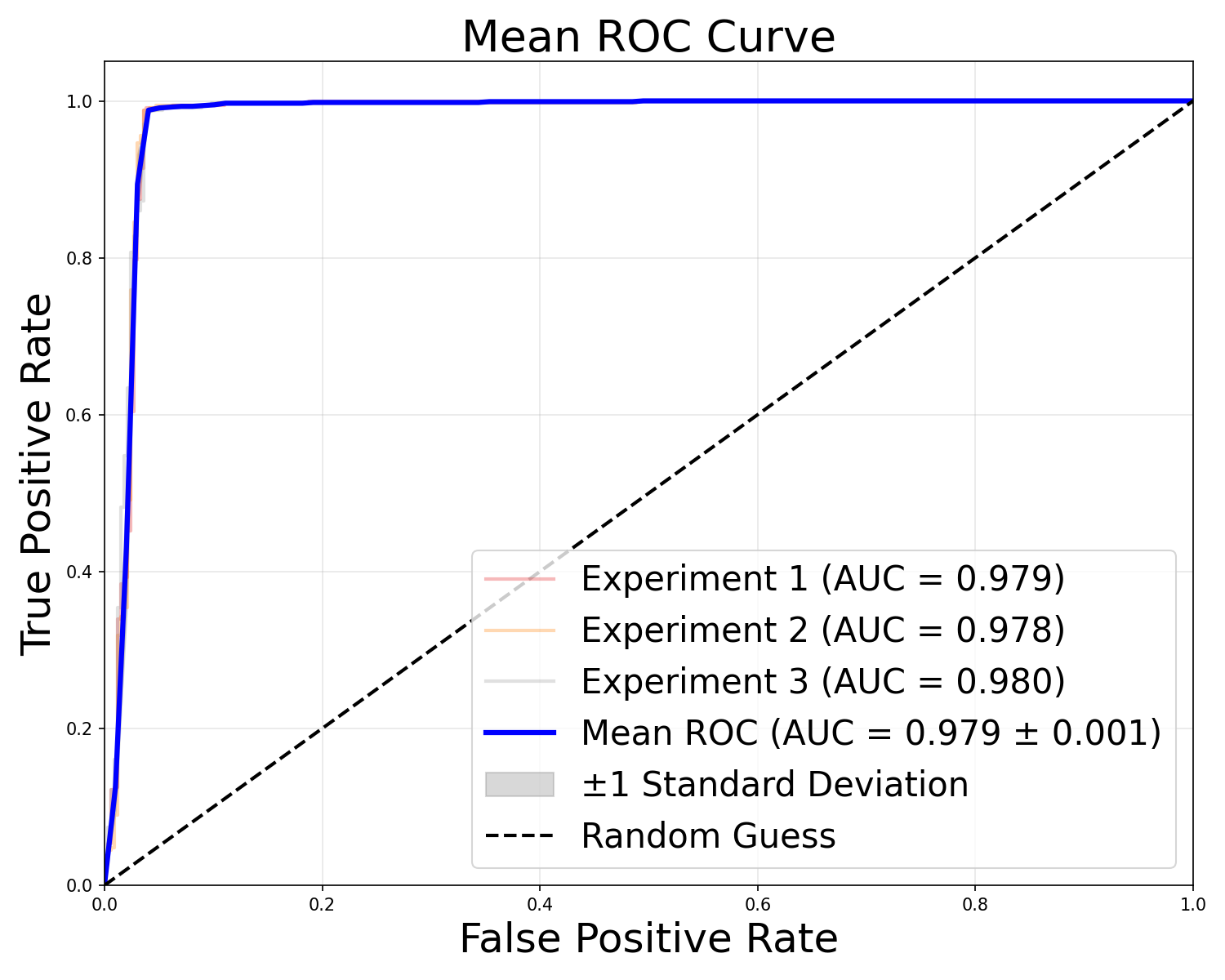}
        \caption{UNIT}
    \end{subfigure}
    \hfill
    \begin{subfigure}[b]{0.246\textwidth}
        \centering
        \includegraphics[width=\textwidth]{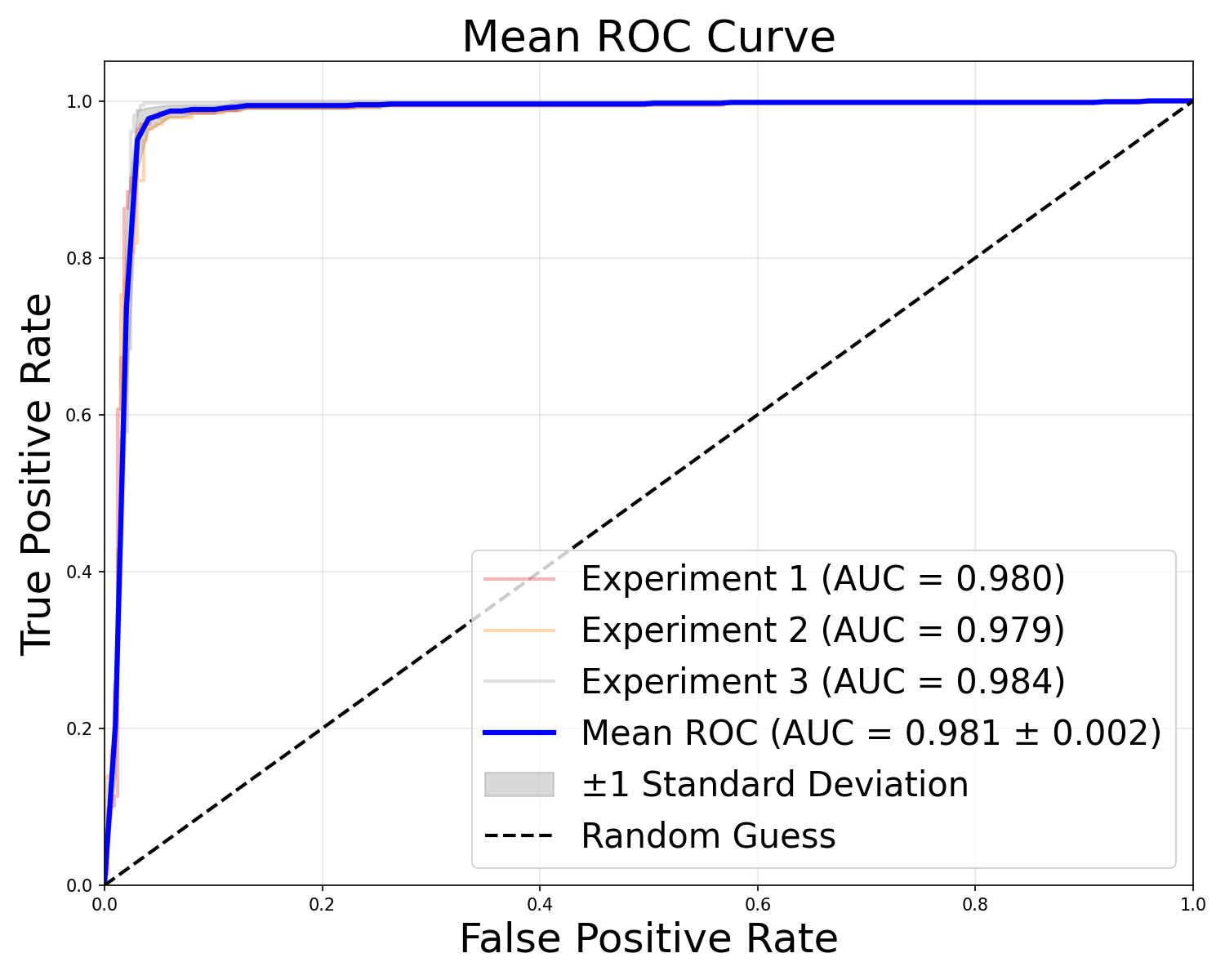}
        \caption{MUNIT}
    \end{subfigure}
    \hfill
    \begin{subfigure}[b]{0.246\textwidth}
        \centering
        \includegraphics[width=\textwidth]{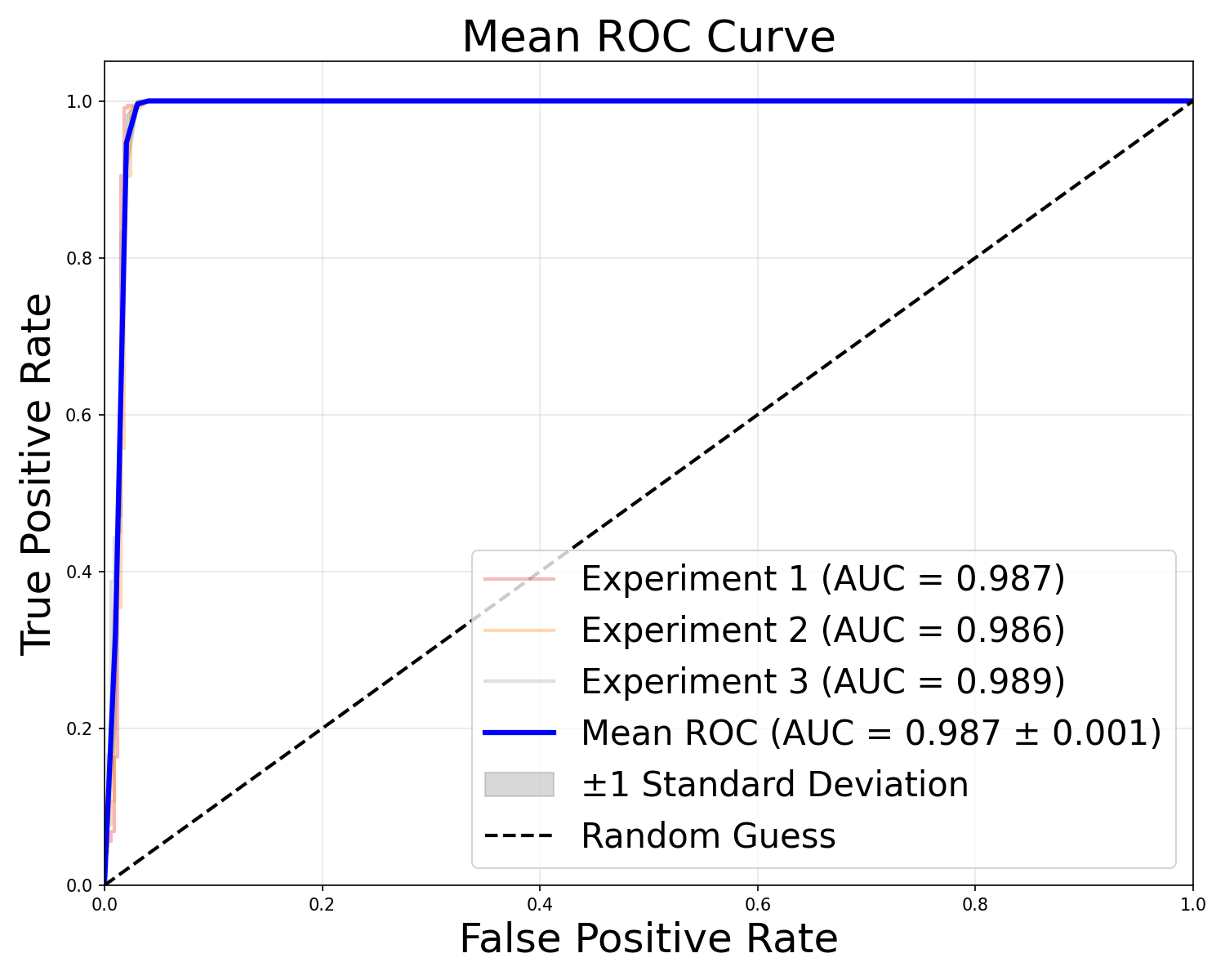}
        \caption{DCLGAN}
    \end{subfigure}

    \caption{Comparison of receiver operating characteristic (ROC) curves among lightweight models for myometrial invasion classification.}
    \label{fig:roc_fourmodels}
\end{figure*}

Figure \ref{fig:roc_fourmodels} visually compares the ROC curves of four lightweight models for classifying myometrial invasion. The proposed LSNet demonstrates the highest overall discriminative ability, achieving the largest area under the curve (AUC), which corresponds to its superior quantitative performance reported in Table \ref{tab:lsnet_comparison}. Compared to MobileNet-V2, EfficientNet, and MobileViT, LSNet maintains a consistently higher true positive rate across all false positive rates, indicating its enhanced capability to distinguish between invasive and non-invasive cases with greater reliability.

\subsection*{Ablation Analysis of LSNet's Core Components}
To quantitatively assess the individual contributions of the two pivotal components in our Lightweight Screening Network (LSNet), namely synthetic data pre-training and gradient-guided sparse attention distillation, we conducted a systematic ablation study. All models in this section were evaluated on the same independent test set used in the main experiment, comprising 336 positive (invasive) and 336 negative (non-invasive) real ultrasound images. The reported results are the mean and standard deviation from three independent runs. The full LSNet (referred to as Full Model) is compared against two ablated versions:

LSNet w/o Syn. Pre-train (No Synthetic Pre-training): This variant is trained from scratch using only the available real ultrasound data, without the initial large-scale pre-training on the hybrid dataset (real + SG-CycleGAN synthetic images). This setup isolates the contribution of the synthetic data augmentation strategy.

LSNet w/o Grad. Distill. (No Gradient Distillation) (Dense Attn.): This version replaces the gradient-guided sparse attention with a standard dense self-attention block, while retaining synthetic pre-training. It specifically evaluates the contribution of the gradient distillation framework to sparsification and knowledge transfer.

\begin{table*}[ht]
\centering
\caption{Ablation Study of the Proposed Lightweight Screening Network (LSNet).}
\label{tab:ablation_study}
{
\renewcommand{\arraystretch}{1.5}
\begin{tabular}{|l|c|c|c|c|c|c|}
\hline
\multirow{2}{*}{Model Variant} & \multirow{2}{*}{Accuracy} & \multirow{2}{*}{Sensitivity} & \multirow{2}{*}{Specificity} & \multirow{2}{*}{Precision} & \multirow{2}{*}{F1-score} & \multirow{2}{*}{ROC AUC} \\
 &  &  &  &  &  &  \\ \hline
LSNet w/o Syn. Pre-train &
\makecell{0.9550 \\ \textcolor{darkgray}{$\pm$ 0.0106}} &
\makecell{0.9467 \\ \textcolor{darkgray}{$\pm$ 0.0246}} &
\makecell{0.9633 \\ \textcolor{darkgray}{$\pm$ 0.0143}} &
\makecell{0.9631 \\ \textcolor{darkgray}{$\pm$ 0.0137}} &
\makecell{0.9545 \\ \textcolor{darkgray}{$\pm$ 0.0111}} &
\makecell{\textbf{0.9914} \\ \textcolor{darkgray}{$\pm$ 0.0017}} \\ \hline

LSNet w/o Grad. Distill. &
\makecell{0.9617 \\ \textcolor{darkgray}{$\pm$ 0.0042}} &
\makecell{0.9700 \\ \textcolor{darkgray}{$\pm$ 0.0071}} &
\makecell{0.9533 \\ \textcolor{darkgray}{$\pm$ 0.0062}} &
\makecell{0.9541 \\ \textcolor{darkgray}{$\pm$ 0.0058}} &
\makecell{0.9620 \\ \textcolor{darkgray}{$\pm$ 0.0043}} &
\makecell{0.9909 \\ \textcolor{darkgray}{$\pm$ 0.0008}} \\ \hline

LSNet (Full Model) &
\makecell{\textbf{0.9836} \\ \textcolor{darkgray}{$\pm$ 0.0024}} &
\makecell{\textbf{0.9950} \\ \textcolor{darkgray}{$\pm$ 0.0014}} &
\makecell{\textbf{0.9722} \\ \textcolor{darkgray}{$\pm$ 0.0051}} &
\makecell{\textbf{0.9729} \\ \textcolor{darkgray}{$\pm$ 0.0048}} &
\makecell{\textbf{0.9838} \\ \textcolor{darkgray}{$\pm$ 0.0024}} &
\makecell{0.9873 \\ \textcolor{darkgray}{$\pm$ 0.0012}} \\ \hline
\end{tabular}
}
\end{table*}

The ablation results in Table \ref{tab:ablation_study} clearly quantify the contribution of each core component. The performance decline of LSNet w/o Syn. Pre-train (accuracy: 0.9550, sensitivity: 0.9467) underscores the critical role of synthetic data. A model trained solely on limited real data tends toward conservative predictions, resulting in insufficient learning for critical minority classes such as deep myometrial invasion. The diverse, structure-preserving synthetic images provided by SG-CycleGAN effectively mitigate class imbalance and enhance the model's generalization and discriminative capabilities. Although LSNet w/o Grad. Distill. retains synthetic data pre-training, its dense attention mechanism lags behind the full model across all performance metrics. This indicates that our approach surpasses mere knowledge transfer. The guided sparsification achieved through gradient distillation enables the student network to focus its computation on regions most critical for the classification decision (e.g., the invasion interface), based on importance signals derived from the teacher model's gradients. Consequently, it yields a more discriminative feature representation while reducing computational cost, ultimately leading to superior sensitivity (0.9950) and overall performance.

\begin{table*}[ht]
\centering
\caption{Ablation Study on Gradient-Guided Sparse Attention with Progressive Sparsity Levels.}
\label{tab:sparsity_ablation}
{
\renewcommand{\arraystretch}{1.5}
\begin{tabular}{|l|c|c|c|c|}
\hline
\textbf{Model Variant} & {\makecell{LSNet w/ Full \\Attention ($k=N$)}} & {\makecell{LSNet w/ Sparse \\Attention \\($k=\lfloor N/10\rfloor$)}} & {\makecell{LSNet w/ Sparse \\Attention\\ ($k=\lfloor N/25\rfloor$)}} & {\makecell{LSNet w/ Sparse \\Attention \\ ($k=\lfloor N/50\rfloor$)}} \\ \hline
{FLOPs} & 0.305G & 0.290G & 0.289G & 0.288G \\ \hline
\makecell{CPU Inference Speed \\(seconds/sample)} & 0.438 s & 0.172 s & 0.157 s & 0.154 s \\ \hline
\makecell{GPU Memory Usage \\(batch size=4)} & 10.1G & 7.9G & 7.8G & 7.8G \\ \hline
{Accuracy} & 0.9821 \textcolor{darkgray}{$\pm$ 0.0024} & 0.9812 \textcolor{darkgray}{$\pm$ 0.0031} & \textbf{0.9836} \textcolor{darkgray}{$\pm$ 0.0024} & 0.9831 \textcolor{darkgray}{$\pm$ 0.0014} \\ \hline
{Sensitivity} & 0.9963 \textcolor{darkgray}{$\pm$ 0.0030} & 0.9940 \textcolor{darkgray}{$\pm$ 0.0049} & 0.9950 \textcolor{darkgray}{$\pm$ 0.0014} & \textbf{0.9970} \textcolor{darkgray}{$\pm$ 0.0024} \\ \hline
{Specificity} & 0.9643 \textcolor{darkgray}{$\pm$ 0.0049} & 0.9683 \textcolor{darkgray}{$\pm$ 0.0014} & \textbf{0.9722} \textcolor{darkgray}{$\pm$ 0.0051} & 0.9692 \textcolor{darkgray}{$\pm$ 0.0014} \\ \hline
{Precision} & 0.9691 \textcolor{darkgray}{$\pm$ 0.0017} & 0.9690 \textcolor{darkgray}{$\pm$ 0.0015} & \textbf{0.9729} \textcolor{darkgray}{$\pm$ 0.0048} & 0.9701 \textcolor{darkgray}{$\pm$ 0.0013} \\ \hline
{F1-score} & 0.9825 \textcolor{darkgray}{$\pm$ 0.0023} & 0.9814 \textcolor{darkgray}{$\pm$ 0.0031} & \textbf{0.9838} \textcolor{darkgray}{$\pm$ 0.0024} & 0.9834 \textcolor{darkgray}{$\pm$ 0.0014} \\ \hline
{ROC AUC} & 0.9905 \textcolor{darkgray}{$\pm$ 0.0013} & \textbf{0.9911} \textcolor{darkgray}{$\pm$ 0.0003} & 0.9873 \textcolor{darkgray}{$\pm$ 0.0012} & 0.9908 \textcolor{darkgray}{$\pm$ 0.0009} \\ \hline
\end{tabular}
}
\end{table*}

Notably, in Table \ref{tab:sparsity_ablation}, even under a more aggressive sparsity setting (LSNet, k=$\lfloor N/50 \rfloor$, and in our experiments, each query token is paired with only the top-8 key-value token pairs as determined by gradient-based importance ranking), the performance degradation remains minimal (Sensitivity: 0.9970, Specificity: 0.9692). This robustness can be attributed to the inherent selectivity of the gradient distillation mechanism. Rather than applying uniform compression, it strategically preserves the inter-token connections with the highest gradient magnitudes, which correspond to those most critical for the classification outcome. Consequently, even with a sparser attention map, the retained features remain highly task-relevant, ensuring reliable diagnostic performance under stringent resource constraints.

\subsection*{Bootstrap Validation Results}

To further evaluate the robustness and statistical reliability of our proposed LSNet, we conducted Bootstrap validation~\cite{liu2019comparison} on the independent test set. For each of the three independent experiments, we performed 10,000 bootstrap resamples on the testing data. This non-parametric resampling approach allows us to estimate the sampling distribution of our performance metrics and compute 95\% confidence intervals, providing a more comprehensive assessment of model stability beyond point estimates.

The Bootstrap validation results across three independent experiments are summarized in Table \ref{tab:bootstrap_validation}. For each performance metric, we report the mean value with standard deviation from bootstrap resampling, along with the 95\% confidence interval (CI).
\begin{table*}[ht]
\centering
\caption{Bootstrap Validation Results for LSNet Across Three Independent Experiments (n=10,000 resamples per experiment)}
\label{tab:bootstrap_validation}
\begin{threeparttable}
\renewcommand{\arraystretch}{1.5}
\resizebox{\linewidth}{!}{
\begin{tabular}{|l|c|c|c|c|c|c|}
\hline
\textbf{Experiment} & {Accuracy} & {Sensitivity} & {Specificity} & {Precision} & {F1-score} & {ROC AUC} \\ \hline

\rowcolor{gray!20}
\multirow{1}{*}{Exp \#1} &
\makecell{0.9866 \\ \textcolor{darkgray}{$\pm$ 0.0044} \\ \textcolor{darkgray}{\small{(0.9777-0.9940)}}} &
\makecell{0.9940 \\ \textcolor{darkgray}{$\pm$ 0.0042} \\ \textcolor{darkgray}{\small{(0.9847-1.0000)}}} &
\makecell{0.9792 \\ \textcolor{darkgray}{$\pm$ 0.0078} \\ \textcolor{darkgray}{\small{(0.9625-0.9938)}}} &
\makecell{0.9795 \\ \textcolor{darkgray}{$\pm$ 0.0077} \\ \textcolor{darkgray}{\small{(0.9632-0.9939)}}} &
\makecell{0.9867 \\ \textcolor{darkgray}{$\pm$ 0.0044} \\ \textcolor{darkgray}{\small{(0.9772-0.9943)}}} &
\makecell{0.9872 \\ \textcolor{darkgray}{$\pm$ 0.0056} \\ \textcolor{darkgray}{\small{(0.9751-0.9969)}}} \\ \hline

\multirow{1}{*}{Exp \#2} &
\makecell{0.9806 \\ \textcolor{darkgray}{$\pm$ 0.0053} \\ \textcolor{darkgray}{\small{(0.9702-0.9896)}}} &
\makecell{0.9941 \\ \textcolor{darkgray}{$\pm$ 0.0042} \\ \textcolor{darkgray}{\small{(0.9848-1.0000)}}} &
\makecell{0.9671 \\ \textcolor{darkgray}{$\pm$ 0.0096} \\ \textcolor{darkgray}{\small{(0.9474-0.9848)}}} &
\makecell{0.9680 \\ \textcolor{darkgray}{$\pm$ 0.0094} \\ \textcolor{darkgray}{\small{(0.9484-0.9853)}}} &
\makecell{0.9808 \\ \textcolor{darkgray}{$\pm$ 0.0052} \\ \textcolor{darkgray}{\small{(0.9699-0.9902)}}} &
\makecell{0.9859 \\ \textcolor{darkgray}{$\pm$ 0.0056} \\ \textcolor{darkgray}{\small{(0.9741-0.9956)}}} \\ \hline

\rowcolor{gray!20}
\multirow{1}{*}{Exp \#3} &
\makecell{0.9836 \\ \textcolor{darkgray}{$\pm$ 0.0049} \\ \textcolor{darkgray}{\small{(0.9732-0.9926)}}} &
\makecell{0.9970 \\ \textcolor{darkgray}{$\pm$ 0.0030} \\ \textcolor{darkgray}{\small{(0.9906-1.0000)}}} &
\makecell{0.9702 \\ \textcolor{darkgray}{$\pm$ 0.0093} \\ \textcolor{darkgray}{\small{(0.9504-0.9874)}}} &
\makecell{0.9710 \\ \textcolor{darkgray}{$\pm$ 0.0090} \\ \textcolor{darkgray}{\small{(0.9518-0.9878)}}} &
\makecell{0.9838 \\ \textcolor{darkgray}{$\pm$ 0.0049} \\ \textcolor{darkgray}{\small{(0.9736-0.9926)}}} &
\makecell{0.9888 \\ \textcolor{darkgray}{$\pm$ 0.0047} \\ \textcolor{darkgray}{\small{(0.9789-0.9969)}}} \\ \hline

{\makecell{Overall \\Mean $\pm$ SD}} &
\makecell{0.9836 \\ \textcolor{darkgray}{$\pm$ 0.0024}} &
\makecell{0.9950 \\ \textcolor{darkgray}{$\pm$ 0.0014}} &
\makecell{0.9722 \\ \textcolor{darkgray}{$\pm$ 0.0051}} &
\makecell{0.9729 \\ \textcolor{darkgray}{$\pm$ 0.0048}} &
\makecell{0.9838 \\ \textcolor{darkgray}{$\pm$ 0.0024}} &
\makecell{0.9873 \\ \textcolor{darkgray}{$\pm$ 0.0012}} \\ \hline
\end{tabular}
}
\begin{tablenotes}
\footnotesize
\item \textit{Note: Values are presented as mean $\pm$ standard deviation with 95\% confidence intervals in parentheses.}  
\end{tablenotes}
\end{threeparttable}
\end{table*}

The Bootstrap validation demonstrates excellent and consistent performance across multiple resampled test sets. The narrow confidence intervals for all key metrics across three independent experiments indicate exceptional discriminative ability and stability of our LSNet. For example, the ROC AUC intervals (Experiment 1: 0.9751-0.9969; Experiment 2: 0.9741-0.9956; Experiment 3: 0.9789-0.9969) remain consistently high and tight. The overall mean accuracy across experiments was 0.9836$\pm$0.0024, accompanied by consistently high sensitivity (0.9950$\pm$0.0014) and specificity (0.9722$\pm$0.0051). These results confirm that the model maintains a balanced performance profile without substantial trade-offs between the detection of invasive cases and the avoidance of false positives.

\begin{figure*}[t]
\centering  
\includegraphics[width=0.99\linewidth]{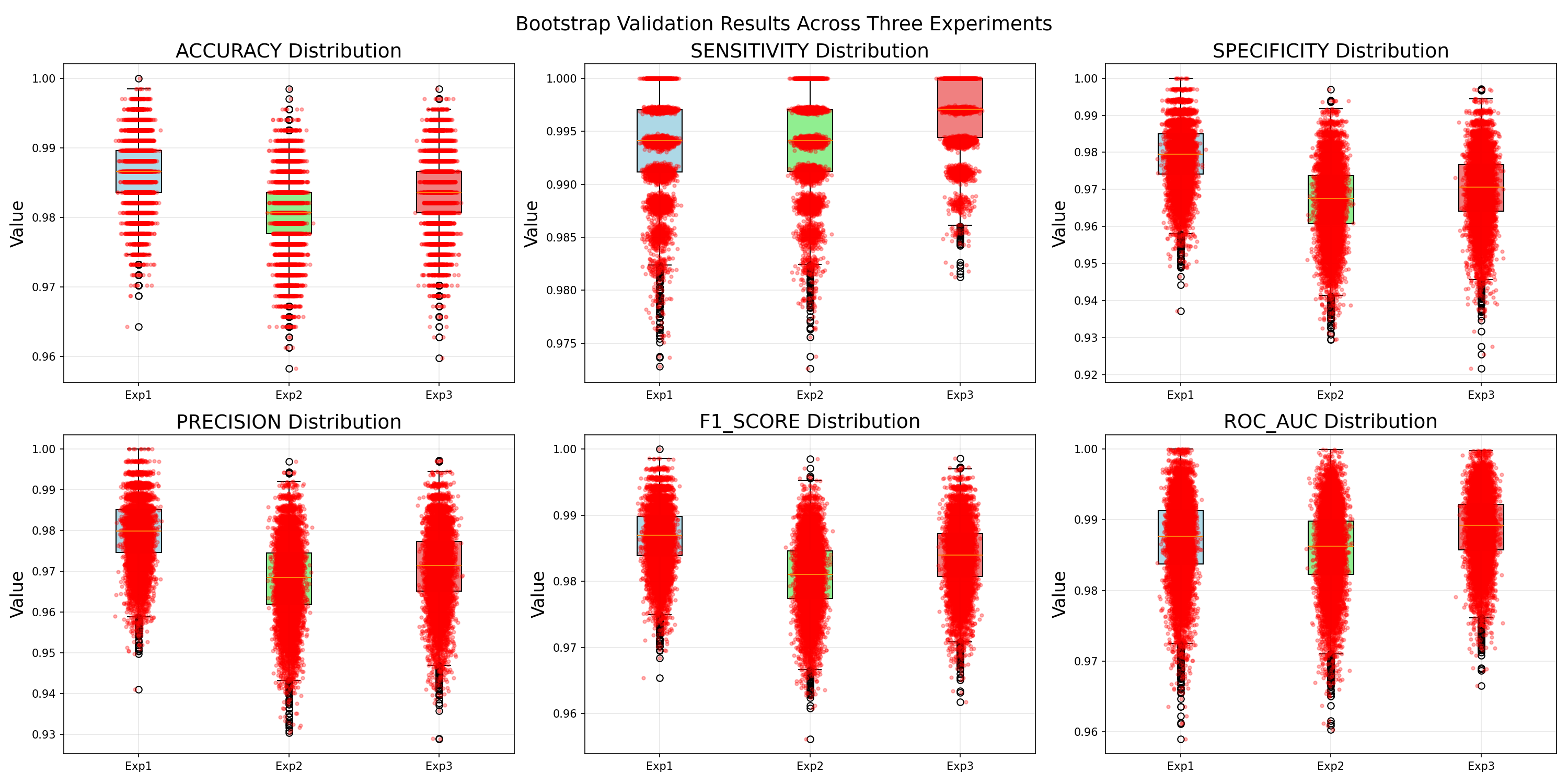}
\caption{Boxplots comparing Bootstrap distributions of performance metrics across three independent experiments. Each box shows the median (center line), interquartile range (box bounds), and range (whiskers) for Accuracy, Sensitivity, Specificity, Precision, F1-score, and ROC AUC.
\label{fig:Boxplots}}
\end{figure*}

Boxplots (Figure \ref{fig:Boxplots}) and histograms (Appendix Figures \ref{fig:bootstrap_validation_results1}-\ref{fig:bootstrap_validation_results3}) also confirm that LSNet maintains tight performance distributions with minimal skewness across all bootstrap resamples.

%
%

\subsection*{Theoretical screening performance}
To assess the potential clinical utility and population-level impact of the proposed Lightweight Screening Network (LSNet), we performed a theoretical screening performance analysis under varying disease prevalence scenarios. This analysis estimates key public health metrics: Positive Predictive Value (PPV), Negative Predictive Value (NPV), Positive Likelihood Ratio (LR+), expected screen-positive rate, and the number needed to screen (NNS) to detect one true positive case, based on the model's validated performance (sensitivity=0.975, specificity=0.955) obtained from the independent test set. The calculation formulas are presented in Appendix B, and the results are summarized in Table \ref{tab:theoreticalscreening}.

\begin{table*}[th]
\centering
\caption{Theoretical screening indicators of the LSNet model (Sensitivity=0.9950, Specificity=0.9722) across populations with different disease prevalences (P).}
\label{tab:theoreticalscreening}
\renewcommand{\arraystretch}{1.5}{
\begin{tabular}{|l|c|c|c|c|c|}
\hline
\textbf{\makecell{Target Population \\\& Prevalence (P)}} & \textbf{PPV} & \textbf{NPV} & \textbf{LR+} & \textbf{\makecell{Expected \\Screen-Positive \\Rate}} & \textbf{\makecell{NNS to Detect\\ One True \\Positive}} \\
\hline
\makecell{General Screening \\(P = 0.02\%)~\cite{hockel2009treatment}} & 0.71\% & $\approx$ 99.99\% & 35.79 & 2.80\% & $\approx$ 5025 \\
\hline
\makecell{Abnormal Postmenopausal \\Bleeding (P = 9\%)~\cite{clarke2018association}} & 78.00\% & 99.95\% & 35.79 & 11.48\% & $\approx$11 \\
\hline
\makecell{Atypical Hyperplasia \\(P = 30.8\%)~\cite{sanni2019348}} & 94.10\% & 99.77\% & 35.79 & 32.57\% & $\approx$3 \\
\hline
\makecell{Lynch Syndrome \\(P = 40\%)~\cite{bonadona2011cancer,ryan2019proportion}} & 96.00\% & 99.66\% & 35.79 & 41.47\% & $\approx$3 \\
\hline
\end{tabular}}
\end{table*}

As shown in Table \ref{tab:theoreticalscreening}, the model's high sensitivity (0.9950) and specificity (0.9722) yield a consistently strong Positive Likelihood Ratio (LR+) of 35.79 across all prevalence settings. This indicates that a positive result from LSNet is approximately 36 times more likely to be observed in a patient with myometrial invasion than in one without, providing substantial diagnostic evidence. The NPV remains excellent ($\geq$99\%) even at high prevalence, ensuring high confidence in ruling out invasion when the model predicts a negative result.

The PPV, however, exhibits significant variation with disease prevalence. In a general screening context with very low prevalence ($P=0.02\%$), the PPV remains modest (0.36\%), reflecting the challenge of identifying rare events and the consequent higher proportion of false positives relative to true positives. In this setting, approximately 5,025 individuals would need to be screened to detect one true case of invasion. In populations with abnormal postmenopausal bleeding ($P=9\%$), the PPV increases to 78.00\% and the NNS drops to about 11. Conversely, in high-prevalence cohorts such as patients with endometrial atypical hyperplasia ($P=30.8\%$) or Lynch syndrome ($P=40\%$), the PPV rises sharply to 94.10\% and 96.00\%, respectively, and the NNS drops dramatically to only 2-3 individuals. This highlights the model's high efficiency and clinical reliability when applied to targeted, high-risk populations.

These theoretical projections underscore the importance of implementing the LSNet within a risk-stratified screening framework. For primary care settings serving a general population, the model can serve as a highly sensitive initial triage tool, identifying a small subset of screen-positive cases for further specialist evaluation or MRI confirmation, thereby optimizing resource allocation. When deployed for surveillance of known high-risk individuals, the model demonstrates robust diagnostic performance, potentially enabling more efficient and accurate monitoring. This analysis confirms that the proposed LSNet is not only computationally efficient but also possesses the predictive characteristics necessary for effective integration into real-world endometrial carcinoma screening pathways.

\subsection*{Comparision with Sonographer Performance}
We selected the testing set for model performance comparison, and the ultrasound images were independently evaluated by 10 sonographers who were unaware of the clinical and pathological results. These sonographers included 5 sonographers with 1-5 years of experience and 5 sonographers with 5-10 years of experience from 4 different medical centres. The subjective assessment of myometrial invasion was performed as follows: Sonographers evaluated the endometrial-myometrial junction on either transverse or longitudinal uterine images. An intact, continuous junction presenting as a distinct hypoechoic band was interpreted as absence of myometrial invasion. In contrast, signs suggestive of tumor breakthrough, such as junctional disruption, irregularity, blurring, or fusion with the myometrial echotexture, were interpreted as presence of myometrial invasion. All judgments were based on the sonographers' subjective evaluation of morphological features in the images.
\begin{table*}[th]
\centering
\caption{Sonographer Performance in Myometrial Invasion Classification.}
\label{tab:Sonographer}
\begin{threeparttable}
\renewcommand{\arraystretch}{1.1}
\begin{tabular}{lccc}
\toprule
\textbf{Sonographer \#} & \textbf{Sensitivity (95\% CI)} & \textbf{Specificity (95\% CI)} & \textbf{ROC AUC (95\% CI)} \\
\midrule
\multicolumn{4}{l}{\textbf{Junior group}} \\
\#1 & 0.69 (0.65-0.73) & 0.83 (0.82-0.84) & 0.759 (0.749-0.770) \\
\#2 & 0.79 (0.76-0.83) & 0.60 (0.59-0.61) & 0.698 (0.687-0.709) \\
\#3 & 0.84 (0.80-0.87) & 0.81 (0.80-0.82) & 0.822 (0.812-0.831) \\
\#4 & 0.65 (0.60-0.69) & 0.80 (0.79-0.81) & 0.725 (0.714-0.736) \\
\#5 & 0.59 (0.55-0.64) & 0.76 (0.75-0.77) & 0.673 (0.662-0.685) \\
\midrule
\multicolumn{4}{l}{\textbf{Senior group}} \\
\#1 & 0.73 (0.69-0.77) & 0.89 (0.89-0.90) & 0.811 (0.802-0.821) \\
\#2 & 0.89 (0.85-0.91) & 0.79 (0.78-0.80) & 0.837 (0.828-0.846) \\
\#3 & 0.74 (0.69-0.78) & 0.81 (0.80-0.82) & 0.772 (0.762-0.782) \\
\#4 & 0.84 (0.80-0.87) & 0.79 (0.78-0.80) & 0.814 (0.805-0.824) \\
\#5 & 0.82 (0.78-0.85) & 0.73 (0.72-0.74) & 0.776 (0.765-0.786) \\
\midrule
Junior, Mean $\pm$ SD & 0.712 $\pm$ 0.102 & 0.760 $\pm$ 0.093 & 0.735 $\pm$ 0.058 \\
Senior, Mean $\pm$ SD & 0.804 $\pm$ 0.068 & 0.802 $\pm$ 0.058 & 0.802 $\pm$ 0.028 \\
Overall, Mean $\pm$ SD & 0.758 $\pm$ 0.095 & 0.781 $\pm$ 0.076 & 0.769 $\pm$ 0.055 \\
\bottomrule
\end{tabular}
\begin{tablenotes}
\footnotesize
\item \textit{Note: AUC, area under the curve; CI, confidence interval.}
\end{tablenotes}
\end{threeparttable}
\end{table*}

As summarized in Table \ref{tab:Sonographer}, the overall diagnostic performance across 10 sonographers yielded a sensitivity of 0.758 $\pm$ 0.095, specificity of 0.781 ($\pm$ 0.076, and ROC AUC of 0.769 $\pm$ 0.055. The senior group demonstrated moderately higher performance (sensitivity: 0.804 $\pm$ 0.068; specificity: 0.802 $\pm$ 0.058; ROC AUC: 0.802 $\pm$ 0.028) compared to the junior group. In stark contrast, the proposed LSNet achieved substantially superior and more consistent performance, with a sensitivity of 0.9950 $\pm$ 0.0014, specificity of 0.9722 $\pm$ 0.0051, and ROC AUC of 0.9873 $\pm$ 0.0012, highlighting its potential to provide more reliable and uniform diagnostic support in primary care settings.

\section*{{Discussion}}
The timely and accurate detection of myometrial invasion in endometrial carcinoma is crucial for guiding clinical management and improving patient outcomes. Ultrasound remains the primary screening modality in primary care due to its accessibility and low cost, yet its diagnostic accuracy is limited by inter-observer variability, subtle imaging features, and severe class imbalance in training data. To address these challenges, we developed an integrated deep learning framework comprising a Structure-Guided Cycle-Consistent Adversarial Network (SG-CycleGAN) for cross-modal image synthesis and a Lightweight Screening Network (LSNet) with Gradient Distillation for efficient invasion classification. Our study demonstrates that this approach generates synthetic ultrasound images that effectively preserve the diagnostically critical anatomical structures from MRI, while also achieving superior classification performance with low computational overhead. This dual capability highlights its potential for practical deployment in resource-constrained settings.

When compared with human sonographers, LSNet exhibited substantially higher and more consistent diagnostic performance (sensitivity 0.9950 vs. 0.758, specificity 0.9722 vs. 0.781, ROC-AUC 0.9873 vs. 0.769). This gap was especially pronounced for junior sonographers, whose performance varied widely. The model's ability to provide objective, reproducible assessments could help reduce diagnostic variability and support less-experienced clinicians in primary care settings. Furthermore, the theoretical screening analysis (Table \ref{tab:theoreticalscreening}) indicated that the model maintains high positive predictive value in high-risk populations (e.g., patients with abnormal postmenopausal bleeding or Lynch syndrome), suggesting its utility in targeted screening programs.

Our work also aligns with the growing emphasis on ethical and practical AI deployment in medicine. To ensure data privacy and security, we recommend operating the LSNet offline in clinical environments, minimizing exposure to external networks. The use of a large, multicentre dataset for training and validation helps mitigate bias and improve generalizability. Continuous monitoring and model updates, overseen by a multidisciplinary ethics committee, will be essential to maintain performance equity and adapt to evolving clinical needs.


\section*{{Methods}}

\subsection*{Unpaired Image-to-Image Translation via Structure-Guided Cycle-Consistent Adversarial Network}
This section presents our Structure-Guided Cycle-Consistent Adversarial Network (SG-CycleGAN) designed to tackle the data imbalance in endometrial carcinoma screening by generating ultrasound images from MRI images. Unlike conventional augmentation techniques~\cite{goceri2023medical,garcea2023data}, our framework performs unpaired translation from MRI to ultrasound, a clinically relevant strategy, as patients with ambiguous ultrasound findings frequently undergo MRI for further evaluation. This approach directly mitigates the severe class imbalance and sample scarcity by synthesizing diverse and pathologically meaningful ultrasound images. 

\subsubsection*{Problem Definition and Preliminaries}
Building upon this clinical context, we formalize the image translation task as follows. Let $\mathcal{X}$ and $\mathcal{Y} $ represent the MRI and ultrasound image domains, respectively. We assume access to unpaired datasets: $ {\mathcal{U}}_{\mathcal{X}} = \{x_1, x_2, ..., x_n\} $ where $ x \in \mathcal{X}  $, and $ \mathcal{U}_{\mathcal{Y}} = \{y_1, y_2, ..., y_m\}  $ where $ y \in \mathcal{Y}  $. Our objective is to learn two mapping functions:

\begin{equation}
\begin{array}{c}  
G: \mathcal{X} \rightarrow \mathcal{Y} : \text{Translates images from domain } \mathcal{X} \text{ to } \mathcal{Y}\\[8pt]
F: \mathcal{Y} \rightarrow \mathcal{X} : \text{Translates images from domain } \mathcal{Y} \text{ to } \mathcal{X}.
\end{array}
\end{equation}

\begin{figure*}[t]
\centering  
\includegraphics[width=0.7\linewidth]{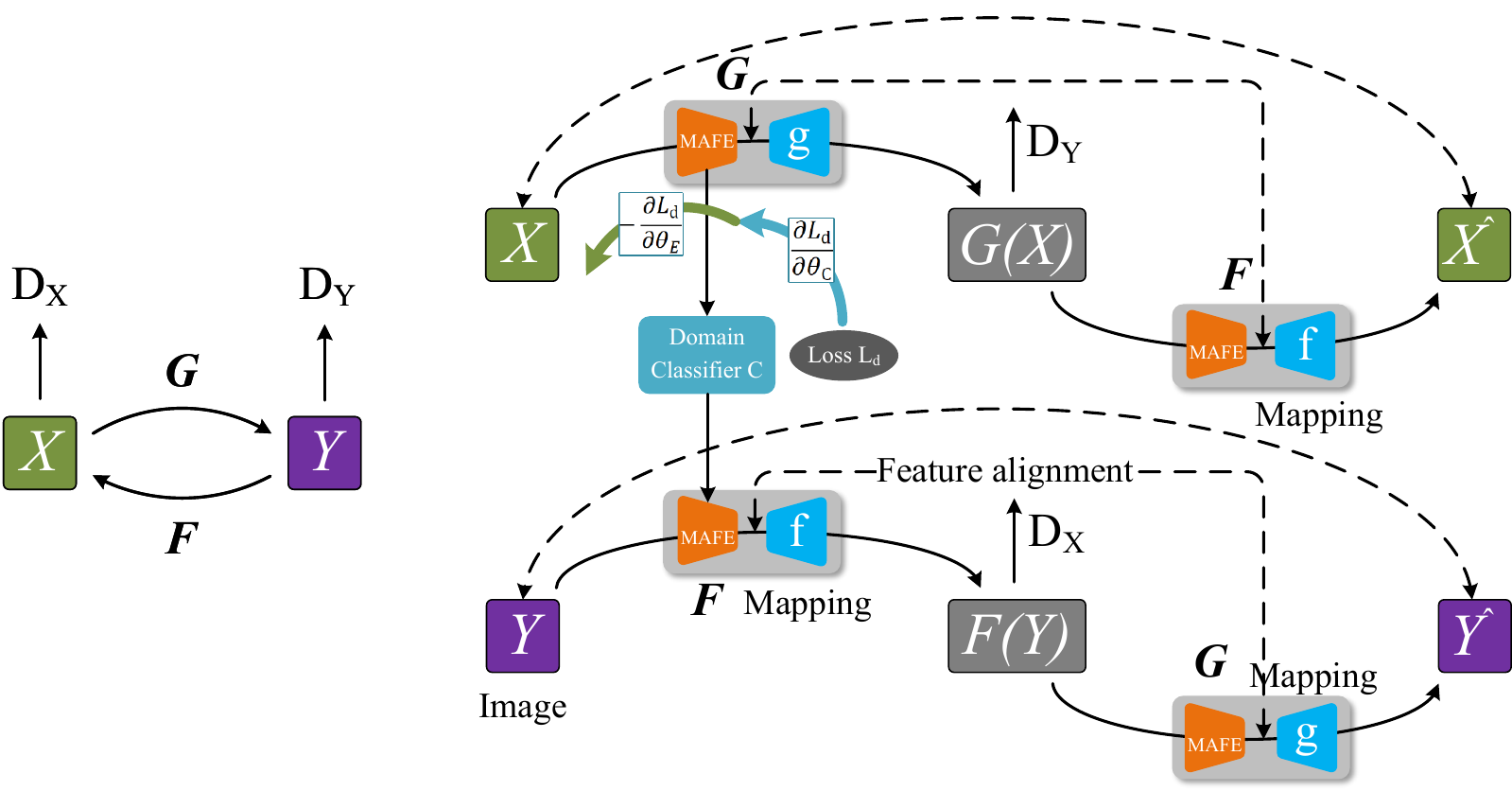}
\\(a)\qquad \qquad\qquad\qquad\qquad\qquad\qquad\qquad(b)\qquad\qquad\qquad\qquad
\caption{(a) CycleGAN Structure; (b) Structure-Guided Cycle-Consistent Adversarial Network (SG-CycleGAN) with Modality-Agnostic Feature Extraction and Gradient reversal
\label{fig:gan}}
\end{figure*}

CycleGAN~\cite{zhu2017unpaired,zhao2020unpaired} addresses unpaired image-to-image translation by learning mappings between two domains. The framework employs two generative adversarial networks (GANs) in a cyclic structure (Figure \ref{fig:gan} (a)): one generator $G$ maps images from $\mathcal{X}$ to $\mathcal{Y} $, while another generator $F$ maps from $\mathcal{Y} $ back to $\mathcal{X}$. Two discriminators, \( D_\mathcal{X} \) and \( D_\mathcal{Y}\), are tasked with distinguishing between real and translated images in their respective domains. The adversarial losses encourage the generated images to be perceptually realistic. Critically, to prevent the generators from producing arbitrary outputs that merely fool the discriminators, CycleGAN introduces a cycle consistency constraint. This constraint mandates that translating an image from one domain to the other and then back again should reconstruct the original image, i.e., \( F(G(x)) \approx x \) and \( G(F(y)) \approx y \). This cycle-consistency loss acts as a powerful regularizer, enforcing semantic preservation during translation and enabling the model to learn meaningful mappings without supervised paired data.

However, when applied to MRI-to-ultrasound translation, the standard CycleGAN framework faces two core challenges. First, it suffers from \textbf{insufficient preservation of high-frequency anatomical details}. The adversarial and cycle-consistency losses ensure global structural coherence but are insensitive to fine textures and sharp boundaries, especially those critical for identifying subtle invasion patterns. This leads to overly smoothed images lacking diagnostic granularity. Second, there is a risk of \textbf{inconsistent anatomical content preservation}. While cycle consistency provides a reconstruction constraint, it lacks direct feature-level or content-level alignment constraints, which cannot enforce that the underlying shared anatomical content (organ shape, position, size, relative relationships) remains unchanged during the modality translation process, as generators may find shortcuts that distort critical regions. Other existing translation models (e.g., UNIT~\cite{liu2017unsupervised}, MUNIT~\cite{huang2018multimodal}, StarGAN v2~\cite{choi2020stargan}) prioritize representational uniformity or style diversity over precise anatomical alignment, rendering them unsuitable for preserving the fine structural details required for EC screening.

Therefore, although these models have achieved success in natural image translation~\cite{nam2024survey}, their core designs are not optimized for the specific requirement of preserving crucial anatomical structural consistency in cross-modal medical image translation.
To address this limitation, we introduce a Modality-Agnostic Feature Extractor (MAFE) as a core intermediary in SG-CycleGAN. MAFE utilizes a gradient reversal layer to actively discard modality-specific appearances, ensuring that the learned feature space encodes only the shared underlying anatomy.

\subsubsection*{Model Architecture}
The SG-CycleGAN framework includes five core components (Figure \ref{fig:gan}(b)):

\textbf{Generator} $ G$: This generator takes an input image from domain $ \mathcal{X}$ (MRI) and produces a translated image in domain $ \mathcal{Y} $ (ultrasound). Its architecture is decomposed into two sub-modules: a modality-agnostic feature extractor (MAFE) and a subsequent mapping function \( g \). The MAFE is responsible for encoding high-level anatomical structures that are common across different imaging modalities.

\textbf{Generator} \( F \): This generator performs the reverse translation, taking an image from domain $ \mathcal{Y} $ (ultrasound) and mapping it to domain $ \mathcal{X}$ (MRI). It shares the identical MAFE sub-module with Generator $ G$, ensuring that both generators build upon a common structural representation. Following the shared MAFE, it uses a dedicated mapping function \( f \) to reconstruct the target modality-specific appearance.

\textbf{Discriminator} \( D_\mathcal{X} \): This component distinguishes between real images from domain $ \mathcal{X}$ and the translated images generated by \( F(\mathcal{Y} ) \), thereby enforcing the visual authenticity of the synthesized MRI images.

\textbf{Discriminator} \( D_\mathcal{Y} \): Similarly, this discriminator differentiates between real images from domain $ \mathcal{Y} $ and the generated outputs from \( G(\mathcal{X}) \), ensuring that the synthetic ultrasound images are perceptually realistic.

\textbf{Domain Classifier} \( C \): The domain classifier \( C \) is tasked with predicting whether an input image originates from domain $ \mathcal{X}$ or $ \mathcal{Y} $. It is applied specifically to the features extracted by the shared MAFE. A gradient reversal layer is employed during training to encourage the MAFE to learn features that are invariant to the image modality, thereby preventing it from encoding domain-specific information and reinforcing its role as a structure-focused feature extractor.

\subsubsection*{Loss Functions and Training Process}
The model is trained with three types of loss: adversarial loss, cycle consistency loss, and optional identity loss.

\textbf{Adversarial Loss}: The adversarial loss enforces translated images to be indistinguishable from real images in the target domain. For generator $ G$ and discriminator \( D_\mathcal{Y} \), the loss is defined as:

\begin{equation}
\begin{aligned}
\mathcal{L}_{GAN}(G, D_\mathcal{Y},&\mathcal{X},\mathcal{Y})=\mathbb{E}_{y \sim \mathcal{U}_\mathcal{Y}} \left[ \log D_\mathcal{Y}(y) \right] \\
&+ \mathbb{E}_{x \sim \mathcal{U}_\mathcal{X}} \left[ \log \left(1 - D_\mathcal{Y}(G(x)) \right) \right].
\end{aligned}
\end{equation}
Similarly, for generator $ F$ and discriminator $ D_\mathcal{X}$:
\begin{equation}
\begin{aligned}
\mathcal{L}_{GAN}(F, D_\mathcal{X},\mathcal{Y}&,\mathcal{X}) =\mathbb{E}_{x \sim \mathcal{U}_\mathcal{X}} \left[ \log D_\mathcal{X}(x) \right] \\
&+ \mathbb{E}_{y \sim \mathcal{U}_\mathcal{Y}} \left[ \log \left(1 - D_\mathcal{X}(F(y)) \right) \right].
\end{aligned}
\end{equation}

\textbf{Cycle and Modality-agnostic Feature Consistency Loss}: To ensure the mappings are reversible, we introduce the cycle consistency loss. To ensure the mappings are reversible and semantically meaningful, we employ the cycle consistency loss, a cornerstone of the CycleGAN framework. This loss mandates that an image translated from one domain to the other and then back to the original domain should be reconstructed faithfully. Formally, for forward cycle $ x \rightarrow G(x) \rightarrow F(G(x))$, the loss requires $F(G(x)) \approx x$,and for backward cycle $y \rightarrow F(y) \rightarrow G(F(y)) $, it requires $ G(F(y)) \approx y $. The cycle loss is:

\begin{equation}
\begin{aligned}
\mathcal{L}_{cyc}(G, F) = &\mathbb{E}_{x \sim \mathcal{U}_\mathcal{X}} \left[ \left\| F(G(x)) - x \right\|_1 \right] \\
&+ \mathbb{E}_{y \sim \mathcal{U}_\mathcal{Y}} \left[ \left\| G(F(y)) - y \right\|_1 \right].
\end{aligned}
\end{equation}

While cycle consistency loss effectively ensures translation reversibility, its pixel-level operation lacks explicit constraints for cross-modal anatomical alignment. To address this, we propose a Modality-agnostic Feature Consistency Loss. By requiring the Modality-Agnostic Feature Extractor (MAFE) to encode the shared anatomical content of both MRI and synthesized ultrasound into identical representations, this loss directly preserves fine-grained structural fidelity during translation.

Formally, given an input MRI image $x_i \in \mathcal{U}_{\mathcal{X}}$ and its synthetic ultrasound counterpart $\hat{u} = G(x_i)$, both are processed by the shared MAFE to extract their respective feature maps, $E(x_i)$ and $E(\hat{u})$. To enforce the aforementioned structural alignment, the Modality-agnostic Feature Consistency Loss minimizes the $\mathcal{L}_1$ distance between these representations:
\begin{equation}
\begin{aligned}
\mathcal{L}_{{maf}}(G,& F)= \mathbb{E}_{x \sim \mathcal{U}_\mathcal{X}} \left[ \| E(F(G(x))) - E(x) \|_1 \right] \\[8pt]
& + \mathbb{E}_{\hat{u} \sim p_{\text{data}}(\hat{u})} \left[ \| E(G(F(\hat{u}))) - E(\hat{u}) \|_1 \right].
\end{aligned}
\end{equation}

By applying this constraint directly in the high-level feature space, the model is explicitly guided to overcome the two key limitations of standard CycleGAN. First, to mitigate the loss of high-frequency details, the feature consistency loss compels the generator to preserve fine-grained structural information, such as the precise boundaries of the endometrium-myometrium junction, throughout the translation process. Second, to ensure consistent anatomical content preservation, this loss explicitly enforces alignment in a modality-agnostic feature space, guaranteeing that the underlying shared anatomical structures remain invariant during the modality translation.

\textbf{Domain Discrimination Loss}: To enforce modality-invariant feature learning in the MAFE, we incorporate a domain discrimination loss coupled with a gradient reversal layer (GRL). The domain classifier \( C \), parameterized by $(W, z)$, takes features extracted by the MAFE and produces a domain prediction through a sigmoid activation:
\begin{equation}
\begin{aligned}
C(E(x); W, z) = \sigma(W^T E(x) + z),
\end{aligned}
\end{equation}
where $\sigma$ denotes the sigmoid function, $E(x)$ represents the feature maps extracted by the MAFE, and $d_i \in \{0,1\}$ is the binary domain label indicating whether sample $x_i$ originates from the source (MRI) or target (ultrasound) domain.

The domain discrimination loss is defined as the binary cross-entropy between the predicted domain probabilities and the true domain labels:
\begin{equation}
\begin{aligned}
\mathcal{L}_d(C,E)= \frac{1}{n}&\sum_{i=1}^n \mathcal{L}_d^i(C(E(x_i)), d_i) \\
                   &\quad + \frac{1}{m}\sum_{i=n+1}^N \mathcal{L}_d^i(C(E(x_i)), d_i), \\
\mathcal{L}_d^i(C(E(x_i)), &d_i)= - \left[ d_i \log(C(E(x_i))) \right. \\
                                & \left. + (1-d_i) \log(1 - C(E(x_i))) \right],
\end{aligned}
\end{equation}
where \( n \) denotes the number of samples in the MRI domain $\mathcal{X}$, and \( m \) denotes the number of samples in the ultrasound domain $\mathcal{Y}$.

During training, the domain classifier parameters $(W, z)$ are optimized to minimize this discrimination loss, thereby improving its ability to distinguish between the two modalities. Simultaneously, the gradient reversal layer reverses the gradient sign during back-propagation through the MAFE, encouraging the feature extractor to learn modality-agnostic representations that confuse the domain classifier.

\textbf{Identity Loss}: To enhance the realism and anatomical fidelity of synthesized ultrasound images, we incorporate an identity loss. This loss constrains the generators to behave as identity mappings when processing images from the target domain. For instance, a real ultrasound image fed into the ultrasound generator should be output unchanged. By preventing the generators from introducing unrealistic artifacts or altering fundamental image properties, the identity loss ensures that the synthesized images not only possess realistic structural outlines but also maintain the authentic appearance of the ultrasound modality, thereby improving their diagnostic relevance:
\begin{equation}
\begin{aligned}
\mathcal{L}_{id}(G, F) =& \mathbb{E}_{y \sim \mathcal{D}_Y} \left[ \left\| G(y) - y \right\|_1 \right] \\
&+ \mathbb{E}_{x \sim \mathcal{D}_X} \left[ \left\| F(x) - x \right\|_1 \right].
\end{aligned}
\end{equation}

\textbf{Total Loss}: The total loss for training the model is a weighted combination of the above losses:
\begin{equation}
\begin{aligned}
\mathcal{L}(G, &F,C, D_X, D_Y) =\mathcal{L}_{GAN}(G, D_Y, \mathcal{X}, \mathcal{Y}) +\\
&\quad \mathcal{L}_{GAN}(F, D_X, \mathcal{Y}, \mathcal{X})+ \lambda (\mathcal{L}_{cyc}(G, F)+ \\
& \mathcal{L}_{{maf}}(G, F)) + \mu ( \mathcal{L}_{d}(C, E) + \mathcal{L}_{id}(G, F) ) ,
\end{aligned}
\end{equation}
where $ \lambda$ (set to 10) controls the strength of the Consistency constraint, and $ \mu$ (set to 5) balances the domain discrimination and identity loss. During training, we minimize the total loss with respect to $ G$,$ F$ and $ C$, and maximize it with respect to $ D_X$ and $ D_Y$.

The training process follows a minimax optimization scheme with alternating updates:

1) Update discriminators $ D_X$ and $ D_Y$ to maximize the adversarial loss, improving their ability to distinguish real images from translated images in their respective domains.

2) Update generators $ G$ and $ F$ to minimize the total loss $\mathcal{L}_{\text{total}}$, focusing on generating realistic images that can fool the discriminators while satisfying cycle consistency, feature consistency, and identity constraints.

3) Update domain classifier $ C$ to minimize the domain discrimination loss $\mathcal{L}_d$, enhancing its capability to distinguish between MRI and ultrasound domains based on the features extracted by the modality-agnostic feature extractor (MAFE). Update MAFE using the reversed gradients from the domain discrimination loss ($-\frac{\partial \mathcal{L}_d}{\partial \theta_E}$), encouraging it to learn modality-invariant features that confuse the domain classifier, thereby enforcing structural consistency across modalities.

4) Repeat until convergence, using Adam optimizer~\cite{adam2014method} with a learning rate of 2e-4 and batch size of $32$.

To ensure training stability, the gradient reversal layer weight $\lambda_{GRL}$ follows a progressive scheduling strategy: $\lambda_{GRL} = \frac{2}{1 + \exp(-\gamma \cdot p)} - 1$, where $\gamma = 5$ controls the scheduling rate and $p = \frac{\text{current epoch}}{\text{total epochs}}$ represents the training progress. This allows the model to initially focus on learning the primary translation task before gradually introducing stronger domain adversarial constraints. The complete training procedure is outlined in Algorithm~\ref{alg:sg_cyclegan} of the Appendix.

\subsection*{Lightweight Screening Network via Gradient Distillation}
\begin{figure*}[t]
\centering  
\includegraphics[width=0.98\linewidth]{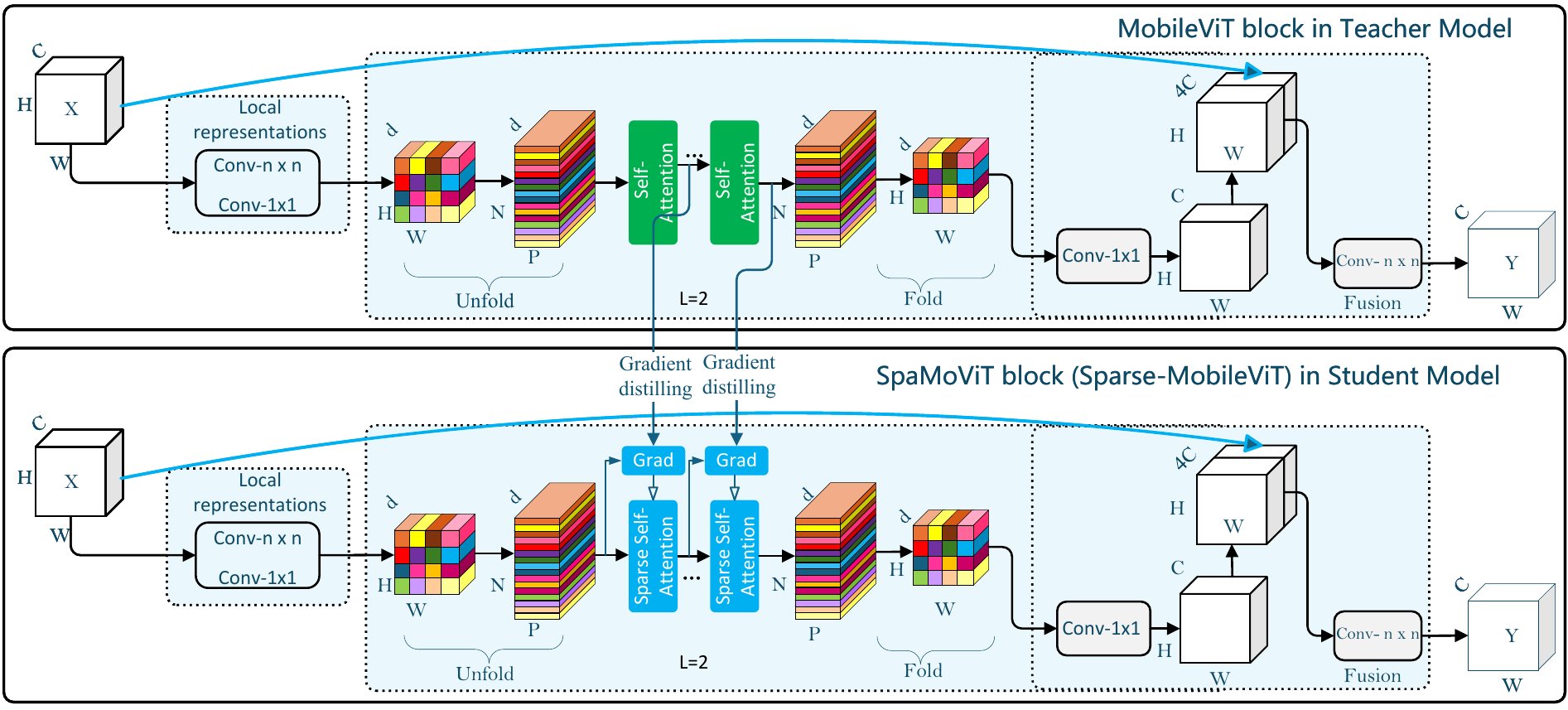}
\caption{Gradient distillation with simulated gradient learning, guiding the student network's sparse attention mechanism.}
\label{fig:distillation}
\end{figure*}

\subsubsection*{Motivation}
Building upon the synthetic data generated by the SG-CycleGAN, this section presents the architectural design of a lightweight screening network. The model is optimized for deployment in resource-constrained primary care environments. Central to this design is a \textbf{Gradient Distillation} framework that serves dual purposes: it facilitates the transfer of discriminative knowledge from a computationally intensive teacher model to a compact student network, while guiding strategic sparsification of the student's attention mechanism.
\subsubsection*{Architecture: Student and Teacher Backbones}
Both the student and teacher networks are constructed based on the MobileViT~\cite{mehta2022mobilevit} architecture, ensuring architectural homogeneity for effective knowledge and gradient distillation. The core of MobileViT lies in its ability to model both local and global dependencies through a hybrid design. A standard MobileViT block transforms an input tensor \( \mathbf{x} \in \mathbb{R}^{H \times W \times C} \) by first applying a local feature extractor (a convolutional block) to produce \( \mathbf{x}_L \). Subsequently, \( \mathbf{x}_L \) is unfolded into \( N \) non-overlapping patches of size \( P \times P \), which are then processed by a transformer block to capture global contexts, yielding \( \mathbf{x}_G \). Finally, \( \mathbf{x}_G \) is folded back to the original spatial dimensions.

The key distinctions between the teacher and student models are twofold: 1) Model Scale: the teacher network employs a larger hidden dimension \( d_{\text{teacher}} \) across all layers, endowing it with superior representational capacity compared to the student network, which uses a compact hidden dimension \( d_{\text{student}} \) (where \( d_{\text{student}} \ll d_{\text{teacher}} \)); 2) Core Building Block: the teacher network utilizes the standard MobileViT block to fully leverage its performance potential. In contrast, the student network incorporates a Sparse MobileViT Block, where the self-attention mechanism is sparsified under the guidance of gradient distillation. This strategic sparsification directs the student's computational focus towards task-critical regions while suppressing irrelevant background and artifacts, thereby enhancing both computational efficiency and feature discriminability.

\subsubsection*{Gradient-Guided Sparse Attention Distillation}
This approach leverages gradient-based importance signals from a teacher model to guide the sparsification of attention in the student network, thereby focusing computational resources on the most diagnostically relevant regions (Figure \ref{fig:distillation}). In the context of myometrial invasion classification, where subtle structural alterations at the endometrium-myometrium junction are often decisive, directing the model's focus to these critical areas is paramount.

While conventional knowledge distillation~\cite{gou2021knowledge,fang2026knowledge} often transfers soft labels~\cite{yang2023knowledge} or intermediate feature maps~\cite{xie2024hierarchical} from teacher to student, such methods do not explicitly guide the student's attention mechanism to prioritize task-critical regions. To address this limitation, we utilize the gradient of attention scores in the teacher model as a principled and robust indicator of token-wise importance. The underlying rationale is that the magnitude of an attention score's gradient reflects its sensitivity to the final classification decision: tokens interconnected by attention weights with larger gradients exert a more significant influence on the model's output. By distilling these gradient-based importance maps, we can direct the student's sparse attention mechanism to prioritize the most influential tokens, thereby simultaneously improving classification accuracy and computational efficiency.

\textbf{Gradient-Based Importance Estimation}: Let \( A_t \in \mathbb{R}^{N \times N} \) denote the self-attention score matrix~\cite{sharma2025novel} from the first attention layer of the MobileViT block in the teacher model, where \( N \) is the number of tokens. The gradient of the classification loss \( \mathcal{L}_{cls} \) with respect to \( A_t \) is computed as:
\begin{equation}
\begin{aligned}
\nabla_{A_t} = \frac{\partial \mathcal{L}_{cls}}{\partial A_t}.
\end{aligned}
\end{equation}
Each element \( \nabla_{A_t}(i,j) \) reflects the sensitivity of the classification outcome to the attention score between token \( i \) and token \( j \). A large absolute value of \( \nabla_{A_t}(i,j) \) indicates that the attention from token \( i \) to token \( j \) is critical for the teacher's decision.

\textbf{Gradient-Guided Sparse Attention with Simulated Gradient Learning}: The student network incorporates a gradient-guided sparse attention mechanism that dynamically focuses computational resources on the most diagnostically relevant regions. This approach employs both explicit gradient guidance during training and learned gradient simulation for efficient standalone inference, creating a unified framework that maintains performance while ensuring computational efficiency.

During training, the student network benefits from a dual-source gradient guidance. The teacher model provides explicit supervision through its computed gradient of attention scores, \( \nabla_{A_t} = \frac{\partial \mathcal{L}_{cls}}{\partial A_t} \), which serves as ground-truth importance signals. Simultaneously, an embedded gradient simulator learns to approximate this gradient distribution directly from the student's input features through a gradient simulation loss:
\begin{equation}
\begin{aligned}
\mathcal{L}_{\text{grad-sim}} = \| \Phi(\nabla_{\text{sim}} )- \Phi(\nabla_{A_t}) \|_2^2,
\end{aligned}
\end{equation}
where the gradient simulator generates \( \nabla_{\text{sim}} \). The simulator employs a lightweight attention-based architecture where the input feature tensor $\mathbf{X} \in \mathbb{R}^{B \times N \times D}$ undergoes head-wise decomposition into $H$ parallel subspaces, followed by depthwise separable convolution~\cite{vidhyalakshmi2026efficient} for efficient spatial feature enhancement. The core gradient simulation is achieved through self-correlation computation:
\begin{equation}
\begin{aligned}
\mathbf{Z} &= \text{BN}(\text{PointwiseConv}(\text{DepthwiseConv}(\mathbf{X}_{\text{conv}})))\\
&\nabla_{\text{sim}}  = \text{SELU}(\mathbf{Z} \mathbf{Z}^\top) \in \mathbb{R}^{B \times H \times N \times N},
\end{aligned}
\end{equation}
where, $\mathbf{X}_{conv}$ denotes the $2D$ convolution-compatible feature map obtained by reshaping X, $\mathbf{Z}$ represents the convolved feature tensor, and SELU~\cite{wang2024improved} denotes the Scaled Exponential Linear Unit activation function, which provides self-normalizing properties to stabilize training.

Rather than direct gradient matching, we employ feature-level distillation that focuses on statistical characteristics rather than raw gradient patterns. The feature extraction network $\Phi(\cdot)$ transforms gradients into compact 128-dimensional representations through spatial compression and convolutional processing, enhancing robustness by filtering noise and irrelevant details, improves stability through natural gradient smoothing, and promotes generalization by learning fundamental gradient characteristics rather than memorizing specific values.

These simulated gradients are then processed to determine sparse attention patterns. First, bidirectional probability distributions are computed using temperature-scaled softmax:
\begin{equation}
\begin{aligned}
P_{\text{pos}} &= \text{Softmax}(\nabla_{sim} / \tau),\\
\quad P_{\text{neg}} &= \text{Softmax}(-\nabla_{sim}  / \tau),
\end{aligned}
\end{equation}
with \( \tau = 0.5 \) controlling distribution sharpness. These distributions are then adaptively fused:
\begin{equation}
\begin{aligned}
\nabla_{\text{norm}} = \alpha \cdot P_{\text{pos}} + (1 - \alpha) \cdot (1 - P_{\text{neg}}),
\end{aligned}
\end{equation}
where \( \alpha = 0.7 \) emphasizes positive attention while maintaining sensitivity to suppression cues. For each query token \( i\), the top-\( k\) key tokens with the highest importance weights are selected, where \( k = \lfloor N/25 \rfloor \) and \( N=400 \) in our experimental settings, forming the sparse attention set \( \mathcal{T}_i \).

The sparse attention matrix \( A_s \) is subsequently computed as:
\begin{equation}
\begin{aligned}
A_s(i,j) =
\begin{cases}
\frac{\exp(Q_i \cdot K_j^\top)}{\sum_{l \in \mathcal{T}_i} \exp(Q_i \cdot K_l^\top)}, & \text{if } j \in \mathcal{T}_i \\
0, & \text{otherwise},
\end{cases}
\end{aligned}
\end{equation}
where \( Q \) and \( K \) are the query and key vectors of the student model, respectively.

\subsubsection*{Overall Optimization Objective}
In contrast to conventional distillation approaches that employ a fixed pre-trained teacher, our framework initializes both teacher and student models randomly and trains them concurrently. This design allows the teacher to continuously refine its feature representations while the student progressively learns from the evolving knowledge through gradient distillation.

We implement a progressive decoupling mechanism that dynamically transitions the student's learning dependency from explicit teacher supervision to internal simulation capability. The overall objective function integrates multiple loss components with time-varying weights:
\begin{equation}
\begin{aligned}
\mathcal{L}_{\text{total}} = \mathcal{L}_{\text{cls}} + \mathcal{L}_{\text{distill}} + \lambda(t) \cdot \mathcal{L}_{\text{grad-sim}},
\end{aligned}
\end{equation}
where \( \mathcal{L}_{\text{cls}} \) denotes the standard classification loss, \( \mathcal{L}_{\text{distill}} \) represents the distillation loss, and $\mathcal{L}_{\text{grad-sim}}$ indicates the gradient feature simulation loss between student-simulated gradients and teacher-provided gradients. $\lambda(t)$ represents training progress. This formulation orchestrates three distinct learning phases: during initial training ($\lambda \approx 1$), strong dependence on teacher gradients ensures stable initialization; the transitional phase ($0 < \lambda < 1$) enables smooth handover as the gradient simulator progressively assumes control; and final deployment ($\lambda \to 0$) achieves complete student autonomy while maintaining distilled knowledge.

To address training instability from evolving teacher gradients, we incorporate gradient smoothing through exponential smoothing techniques. The smoothed gradient target is computed using an exponential moving average approach:
\begin{equation}
\begin{aligned}
\nabla_{\text{smooth}}^{(t)} = \alpha \cdot \nabla_{\text{smooth}}^{(t-1)} + (1-\alpha) \cdot \nabla_{A_t}^{(t)},
\end{aligned}
\end{equation}
where $\alpha$ controls the smoothing factor, placing greater emphasis on recent gradient patterns while maintaining historical context. This method effectively reduces high-frequency fluctuations while preserving essential learning directions, with the smoothing intensity adaptively tuned throughout training. The Equation (12) then evolves to:
\begin{equation}
\begin{aligned}
\mathcal{L}_{\text{grad-sim}} = \| \Phi(\nabla_{\text{sim}} )- \Phi(\nabla_{smooth}) \|_2^2.
\end{aligned}
\end{equation}


\section*{Data availability}
Access to the data can be granted upon reasonable request and with appropriate approvals from the relevant ethics committee and the affiliated hospitals.

\section*{Code availability}
The source code used in this study can be accessed at: \href{https://github.com/dongjingshan2010/SG-CycleGAN-LSNet}{SG-CycleGAN\&LSNet}.

\section*{References}
\bibliographystyle{vancouver}
\bibliography{lancet-sample}

\subsection*{Acknowledgments}
This work was supported by Noncommunicable Chronic Diseases-National Science and Technology Major Project (Grant No. 2024ZD0537707), National Natural Science Foundation of China (Grant Nos. U23A20398, 82272077), Sichuan Science and Technology Program (Grant No. 2026YFHZ0129), Luzhou Science and Technology Program (Grant No. 2025TXD008) and Research Start-up Foundation of Southwest Medical University (Grant Nos. 00040155).

\subsection*{Contributors}
Chunxiang Zhang conceptualised the study.
Yamei Luo, Jiqing Xuan, Lu Huang, Jin Li, Mengchu Yang and Zeyu Chen handled data acquisition and quality control.
Lu Huang and Jin Li conducted the literature search.
Dongjing Shan, Yong Tang and Chunxiang Zhang drafted the manuscript.
Jiqing Xuan, Mengchu Yang and Fajin Lv performed data analysis and interpreted the data. Dongjing Shan and Chunxiang Zhang contributed to model development and validation.
All authors had access to the raw datasets. The corresponding authors
(Yong Tang and Chunxiang Zhang) verified the data and made the final decision to submit the manuscript for publication. All authors reviewed and approved the final
manuscript.

\subsection*{Declaration of interests}
We declare no competing interests.

\appendix


\subsection*{Appendix A: Algorithm}
\label{AppendixA}
\begin{algorithm}[H]
\caption{Training Procedure for SG-CycleGAN}
\label{alg:sg_cyclegan}
\begin{algorithmic}[1]
\REQUIRE Unpaired MRI dataset $\mathcal{X} = \{x_i\}_{i=1}^N$; ultrasound dataset $\mathcal{Y} = \{y_j\}_{j=1}^M$
\REQUIRE Loss weights: $\lambda_{cyc}=10$, $\lambda_{id}=5$, $\lambda_d=5$
\REQUIRE Optimizer: Adam with learning rate $2\times10^{-4}$, batch size $32$
\ENSURE Trained generators $G$, $F$; discriminators $D_{\mathcal{X}}$, $D_{\mathcal{Y}}$; domain classifier $C$

\STATE Initialize parameters for $G$, $F$, $D_{\mathcal{X}}$, $D_{\mathcal{Y}}$, $C$
\STATE Initialize gradient reversal layer weight $\lambda_{GRL}$ with progressive scheduling: $\lambda_{GRL} = \frac{2}{1+\exp(-\gamma \cdot p)}-1$
\WHILE{not converged}
    \STATE \textbf{Step 1: Update discriminators}
    \FOR{$k$ steps (typically $k=1$)}
        \STATE Sample minibatch $\{x_i\} \subset \mathcal{X}$, $\{y_j\} \subset \mathcal{Y}$
        \STATE Compute adversarial loss $\mathcal{L}_{\text{GAN}}$ for $D_{\mathcal{Y}}$ and $D_{\mathcal{X}}$
        \STATE Update $D_{\mathcal{Y}}$ to maximize $\mathcal{L}_{\text{GAN}}(G, D_{\mathcal{Y}}, \mathcal{X}, \mathcal{Y})$
        \STATE Update $D_{\mathcal{X}}$ to maximize $\mathcal{L}_{\text{GAN}}(F, D_{\mathcal{X}}, \mathcal{Y}, \mathcal{X})$
    \ENDFOR

    \STATE \textbf{Step 2: Update generators}
    \STATE Sample minibatch $\{x_i\} \subset \mathcal{X}$, $\{y_j\} \subset \mathcal{Y}$
    \STATE Compute total generator loss:
    \STATE $\quad \mathcal{L}_{\text{total}} = \mathcal{L}_{\text{GAN}}(G, D_{\mathcal{Y}}, \mathcal{X}, \mathcal{Y}) + \mathcal{L}_{\text{GAN}}(F, D_{\mathcal{X}}, \mathcal{Y}, \mathcal{X})$
    \STATE $\quad + \lambda_{cyc} \cdot \mathcal{L}_{\text{cyc}}(G, F) + \lambda_{id} \cdot \mathcal{L}_{\text{id}}(G, F)$
    \STATE $\quad + \mathcal{L}_{\text{feat}}(G, F, \text{MAFE})$
    \STATE Update $G$ and $F$ to minimize $\mathcal{L}_{\text{total}}$

    \STATE \textbf{Step 3: Update domain classifier and MAFE}
    \STATE Sample minibatch $\{x_i\} \subset \mathcal{X}$, $\{y_j\} \subset \mathcal{Y}$
    \STATE Extract features using MAFE: $f_x = \text{MAFE}(x)$, $f_y = \text{MAFE}(y)$
    \STATE Compute domain discrimination loss $\mathcal{L}_d(C, f_x, f_y)$
    \STATE Update $C$ to minimize $\mathcal{L}_d$
    \STATE Update MAFE using reversed gradients: $-\lambda_{GRL} \cdot \frac{\partial \mathcal{L}_d}{\partial \theta_{\text{MAFE}}}$

    \STATE Update training progress $p$ and adjust $\lambda_{GRL}$
\ENDWHILE
\end{algorithmic}
\end{algorithm}

\subsection*{Appendix B: Visualization of Bootstrap Distributions}

To provide detailed visual evidence of the statistical distribution of model performance, the appendix presents histograms of the Bootstrap validation results from three independent experiments (Appendix Figures B1-B3). Each figure displays the frequency distributions of six key performance metrics across 10,000 resampling iterations, offering further validation of the robustness and reproducibility of the proposed Lightweight Screening Network (LSNet) in terms of distribution shape and stability.

\begin{figure*}[t]
\centering  
\includegraphics[width=0.99\linewidth]{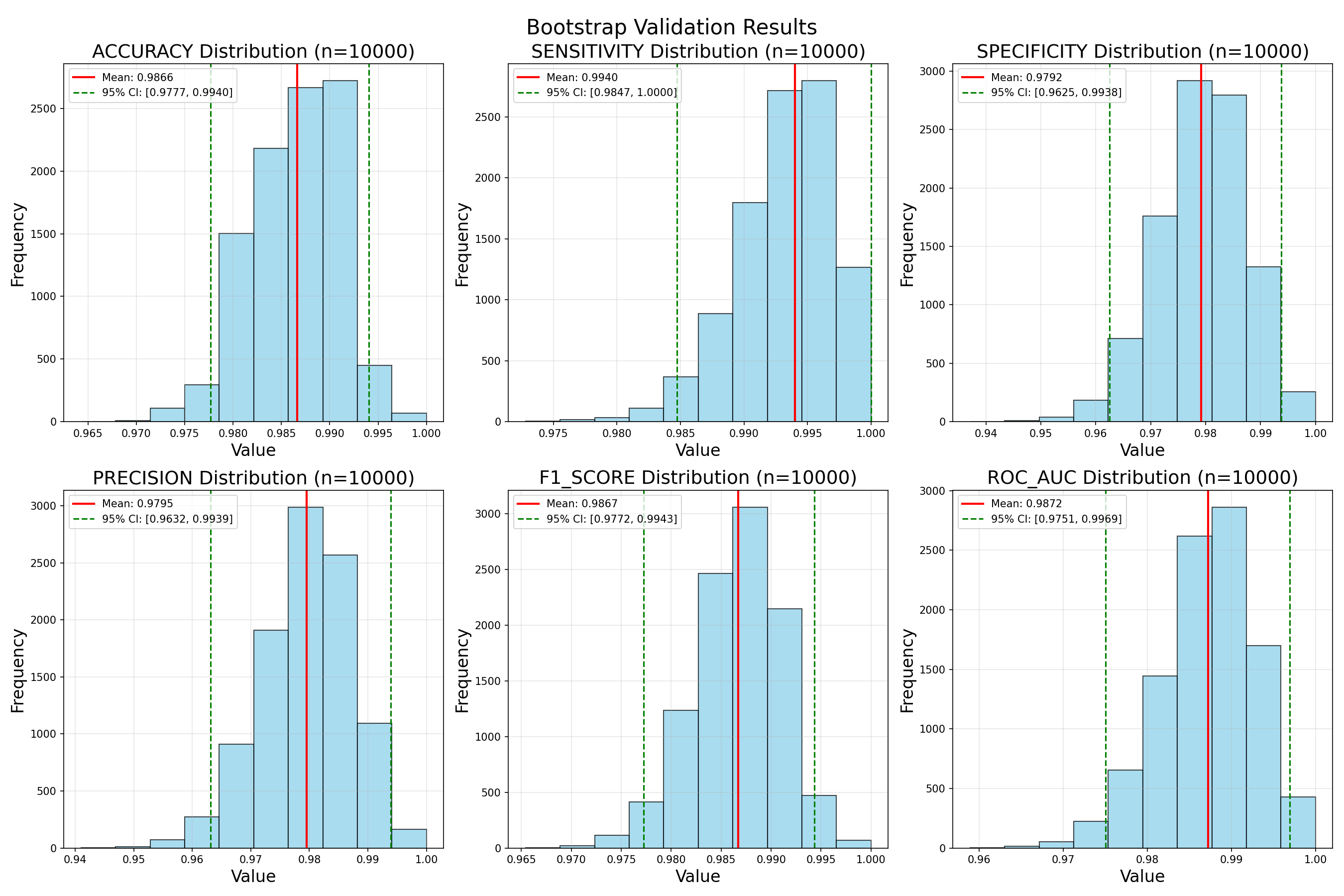}
\caption{Histograms of bootstrap distributions for Experiment 1. The figure displays the frequency distributions of six performance metrics (Accuracy, Sensitivity, Specificity, Precision, F1-score, and ROC AUC) across 10,000 bootstrap resamples. The approximately normal shapes and minimal skewness of the distributions confirm the statistical reliability of the model's performance in the first independent experiment.
\label{fig:bootstrap_validation_results1}}
\end{figure*}

\begin{figure*}[t]
\centering  
\includegraphics[width=0.9\linewidth]{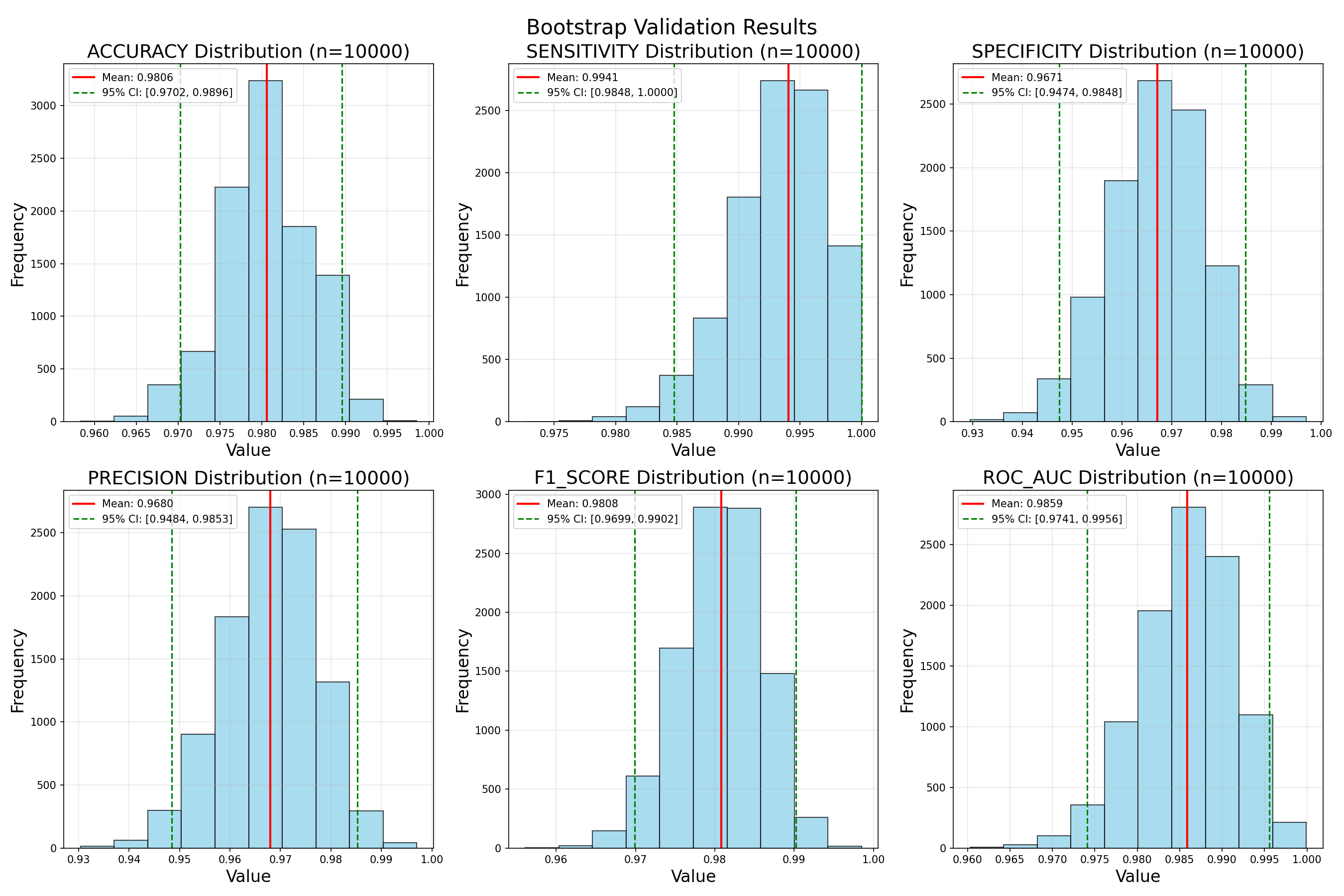}
\caption{Histograms of bootstrap distributions for Experiment 2.
The symmetrical, near-normal shapes of the histograms further validate the stability and reproducibility of the model's results in the second independent trial.
\label{fig:bootstrap_validation_results2}}
\end{figure*}

\begin{figure*}[t]
\centering  
\includegraphics[width=0.9\linewidth]{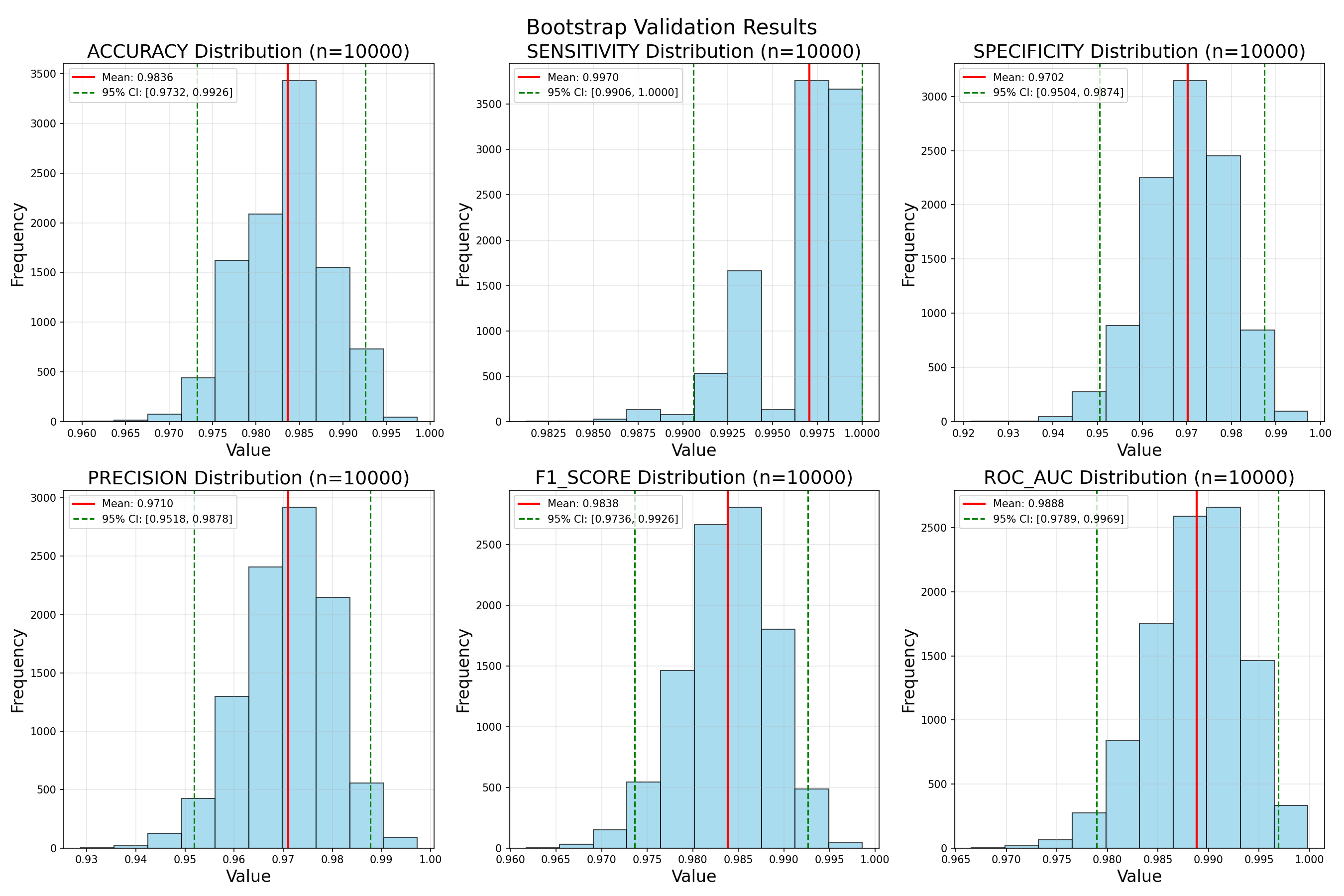}
\caption{Histograms of bootstrap distributions for Experiment 3.
The consistent, well-formed distributions with negligible skew provide additional evidence for the robustness of the proposed method.
\label{fig:bootstrap_validation_results3}}
\end{figure*}

\subsection*{Appendix C: Formulas for Theoretical Screening Performance Metrics}
This appendix outlines the general formulas used to derive the theoretical screening performance metrics presented in Table 7 (Section 3.1.5). The calculations are based on established epidemiological principles linking test performance to population disease prevalence.

\subsubsection*{Parameter Definitions}
\noindent{Let:}

P = Disease prevalence in the target population;

Se = Sensitivity of the screening model;

Sp = Specificity of the screening model;

\subsubsection*{Key Metric Formulas}

Positive Likelihood Ratio (LR+): \[    LR+ = \frac{Se}{1 - Sp} \]

Expected Screen-Positive Rate: \[    \text{Rate} = \left[P \times Se\right] + \left[(1 - P) \times (1 - Sp)\right]    \]

Positive Predictive Value (PPV): \[PPV = \frac{P \times Se}{P \times Se + (1-P)\times (1-Sp)}\]

Negative Predictive Value (NPV): \[
    NPV = \frac{(1 - P) \times Sp}{\left[(1 - P) \times Sp\right] + \left[P \times (1 - Se)\right]}
    \]

Number Needed to Screen (NNS) to Detect One True Positive Case: \[  NNS = \frac{1}{P \times Se}    \]

\end{multicols}
\end{document}